\PassOptionsToPackage{unicode}{hyperref}
\PassOptionsToPackage{hyphens}{url}
\PassOptionsToPackage{dvipsnames,svgnames,x11names}{xcolor}
\documentclass[
  11pt,
]{article}
\usepackage{amsmath,amssymb}
\usepackage{iftex}
\ifPDFTeX
  \usepackage[T1]{fontenc}
  \usepackage[utf8]{inputenc}
  \usepackage{textcomp} 
\else 
  \usepackage{unicode-math} 
  \defaultfontfeatures{Scale=MatchLowercase}
  \defaultfontfeatures[\rmfamily]{Ligatures=TeX,Scale=1}
\fi
\usepackage{lmodern}
\ifPDFTeX\else
\fi
\IfFileExists{upquote.sty}{\usepackage{upquote}}{}
\IfFileExists{microtype.sty}{
  \usepackage[]{microtype}
  \UseMicrotypeSet[protrusion]{basicmath} 
}{}
\makeatletter
\@ifundefined{KOMAClassName}{
  \IfFileExists{parskip.sty}{%
    \usepackage{parskip}
  }{
    \setlength{\parindent}{0pt}
    \setlength{\parskip}{6pt plus 2pt minus 1pt}}
}{
  \KOMAoptions{parskip=half}}
\makeatother
\usepackage{xcolor}
\usepackage[margin=1in]{geometry}
\usepackage{longtable,booktabs,array}
\usepackage{calc} 
\usepackage{etoolbox}
\makeatletter
\patchcmd\longtable{\par}{\if@noskipsec\mbox{}\fi\par}{}{}
\makeatother
\IfFileExists{footnotehyper.sty}{\usepackage{footnotehyper}}{\usepackage{footnote}}
\makesavenoteenv{longtable}
\usepackage{graphicx}
\makeatletter
\def\maxwidth{\ifdim\Gin@nat@width>\linewidth\linewidth\else\Gin@nat@width\fi}
\def\maxheight{\ifdim\Gin@nat@height>\textheight\textheight\else\Gin@nat@height\fi}
\makeatother
\setkeys{Gin}{width=\maxwidth,height=\maxheight,keepaspectratio}
\makeatletter
\def\fps@figure{htbp}
\makeatother
\providecommand{\tightlist}{%
  \setlength{\itemsep}{0pt}\setlength{\parskip}{0pt}}
\ifLuaTeX
  \usepackage{selnolig}  
\fi
\IfFileExists{bookmark.sty}{\usepackage{bookmark}}{\usepackage{hyperref}}
\IfFileExists{xurl.sty}{\usepackage{xurl}}{} 
\hypersetup{
  colorlinks=true,
  linkcolor={blue},
  filecolor={Maroon},
  citecolor={Blue},
  urlcolor={blue},
  pdfcreator={LaTeX via pandoc}}

\author{}
\date{}

\begin{document}

{
\hypersetup{linkcolor=}
\setcounter{tocdepth}{2}
\tableofcontents
}
\hypertarget{k-way-energy-probes-for-metacognition-reduce-to-softmax-in-discriminative-predictive-coding-networks}{%
\section{K-Way Energy Probes for Metacognition Reduce to Softmax in
Discriminative Predictive Coding
Networks}\label{k-way-energy-probes-for-metacognition-reduce-to-softmax-in-discriminative-predictive-coding-networks}}

\textbf{JP Cacioli} \emph{Independent Researcher, Melbourne, Australia}
\emph{ORCID: 0009-0000-7054-2014}

\textbf{Date:} 12 April 2026 \textbf{Status:} arXiv preprint, draft

\begin{center}\rule{0.5\linewidth}{0.5pt}\end{center}

\hypertarget{abstract}{%
\subsection{Abstract}\label{abstract}}

We present this as a negative result with an explanatory mechanism, not
as a formal upper bound.

A growing body of work on metacognition in neural networks has
identified failure modes of single-point confidence probes on
transformer-based models. These probes include softmax margins, learned
linear readouts, and post-hoc calibration. The failures motivate
interest in architectures whose internal dynamics might support
categorically different probing methodologies. Predictive coding
networks (PCNs) are an attractive candidate. They are explicitly
energy-based, maintain prediction errors at every layer, and admit a
K-way energy probe in which each candidate class is fixed as a target,
inference is run to settling, and the per-hypothesis settled energies
are compared. The K-way energy probe appears to read a richer signal
source than softmax, because the per-hypothesis energy depends on the
entire generative chain rather than only the output layer.

We argue, both theoretically and empirically, that this appearance is
misleading under the standard Pinchetti-style discriminative PC
formulation. We present an approximate reduction showing that with
target-clamped CE-energy training and effectively-feedforward latent
dynamics, the K-way energy margin decomposes into a monotone function of
the log-softmax margin plus a residual arising from generative-chain
propagation of the clamped target. The decomposition predicts that the
structural probe should track softmax from below, with the residual
contributing a perturbation that is not trained to correlate with
correctness. We do not claim a formal upper bound. We claim that the
decomposition predicts degradation rather than improvement, and we test
this prediction empirically.

We test the reduction across six conditions on CIFAR-10. These include
extended deterministic training, a direct measurement of latent movement
during inference, a post-hoc decoder fairness control on a
backpropagation network, a matched-budget PC versus BP comparison,
test-time Langevin inference across a five-point temperature sweep, and
trajectory-integrated MCPC training in the Oliviers et al.~style. In
every condition the K-way energy probe sat below softmax on the same
network. The gap was stable across training procedures within the
discriminative PC family. Two qualitatively different training regimes,
final-state weight updates and trajectory-integrated MCPC, produced
probes whose AUROC$_2$ values differed by less than \(10^{-3}\) at
deterministic evaluation. This is consistent with the reduction's
prediction that the probe ceiling depends on the energy decomposition
rather than on which member of the discriminative PC training family
produced the weights.

The empirical regime is small. All experiments use a single seed on a
2.1M-parameter network, evaluated on 1280 test images. We do not claim
formal statistical indistinguishability. We frame the result as a
preprint inviting replication at larger scale and across multiple seeds.
We also discuss conditions under which the decomposition would not
apply. These include bidirectional PC, prospective configuration,
generative PC at test time, and energy formulations without CE at the
output. We also suggest directions for productive structural probing
that the present analysis does not foreclose.

\begin{center}\rule{0.5\linewidth}{0.5pt}\end{center}

\hypertarget{introduction}{%
\subsection{1. Introduction}\label{introduction}}

\hypertarget{the-metacognitive-measurement-problem}{%
\subsubsection{1.1 The metacognitive measurement
problem}\label{the-metacognitive-measurement-problem}}

A neural network's metacognition is its ability to estimate its own
likelihood of being correct. Type-2 signal detection theory (Fleming \&
Lau, 2014; Maniscalco \& Lau, 2012) operationalises this by measuring
how well a confidence signal discriminates the network's correct from
incorrect responses, independent of overall accuracy. The standard
metric is Type-2 AUROC (AUROC$_2$). The standard normalisation is the
M-ratio, meta-d$'$/d$'$, which expresses metacognitive sensitivity as a
fraction of Type-1 sensitivity.

Prior work on metacognition in large language models has identified
persistent failure modes of single-point confidence probes. Softmax
margins, entropy of the output distribution, learned linear readouts,
and post-hoc calibration techniques all exhibit these failures. A subset
of frontier models produce confidence signals that are uninformative or
anti-informative for Type-2 discrimination (Cacioli, 2026a, 2026b). One
mechanism that has been proposed for these failures is that
reinforcement learning from human feedback (RLHF) constrains output
behaviour at the readout layer in ways that dominate the underlying
model's internal uncertainty. These findings motivate a methodological
question. Is there a class of confidence probes that is structurally
harder to dominate by output-layer training pathologies?

One natural candidate is \emph{structural probing}. Instead of reading
confidence from the softmax of a trained classifier, read it from a
quantity that depends on multiple architectural components. Such a
quantity is not easily reshaped by output-layer optimisation alone.
Predictive coding networks (PCNs; Rao \& Ballard, 1999; Whittington \&
Bogacz, 2017; Bogacz, 2017; Millidge et al., 2022; Pinchetti et al.,
2024) admit one such quantity, which we call the \emph{K-way energy
probe}. For each candidate class \(k\), fix the output latent at the
one-hot encoding of \(k\), run iterative inference to settling, and read
the settled total energy \(E_k\). The structural prediction is
\(\arg\min_k E_k\). The structural confidence margin is the energy gap
between the second-lowest and lowest settled energies. The probe depends
on the entire generative chain, meaning the top-down prediction of each
layer's latent state from the layer above. The per-hypothesis energy is
computed over all layer-wise prediction errors plus the cross-entropy
term at the output. On the face of it, this is not the same quantity as
softmax.

This paper asks a simple question. Does the K-way energy probe on
standard discriminative PCNs carry metacognitive signal beyond what
softmax on the same network provides, or is the apparent richness
illusory?

\hypertarget{the-answer}{%
\subsubsection{1.2 The answer}\label{the-answer}}

This paper argues that the appearance is illusory. We present an
approximate decomposition. Under standard target-clamped CE-energy
discriminative PC training, with the effectively-feedforward latent
dynamics characteristic of the Pinchetti-style implementation, the K-way
energy margin decomposes as

\[M_k(x) \approx \big[\text{log-softmax margin}\big]_k + R_k(x)\]

The first term is a monotone function of the standard softmax confidence
at hypothesis \(k\). The second term, \(R_k(x)\), is a residual arising
from how the clamped output one-hot propagates through the generative
chain. The residual depends on the trained generative weights and the
input, but is not trained to correlate with correctness. AUROC and
accuracy are invariant to monotone transformations of the confidence
signal. Under this decomposition the K-way energy probe inherits the
log-softmax margin's signal and adds the residual difference as a
perturbation that is not trained to align with correctness. The
decomposition predicts that the probe should track softmax from below
rather than exceed it, with the gap depending on the magnitude of
\(R_k\) relative to the log-softmax margin. We do not claim a formal
upper bound. We claim that the decomposition predicts degradation rather
than improvement, and we test this prediction empirically.

We test the decomposition across six conditions on CIFAR-10 with a
TinyConvPCN architecture (\textasciitilde2.1M parameters):

\begin{enumerate}
\def\labelenumi{\arabic{enumi}.}
\tightlist
\item
  Standard deterministic discriminative PC training extended to 25
  epochs. The structural-probe-vs-softmax gap does not close with
  training. It fluctuates non-monotonically in the 0.066--0.155 AUROC$_2$
  range across checkpoints.
\item
  A direct measurement of latent movement during inference. Standard
  discriminative PC inference is effectively a no-op, with mean
  per-element latent movement on the order of \(10^{-4}\) over 13
  inference steps. This justifies the ``effectively feedforward''
  assumption of the decomposition.
\item
  A backpropagation network with a post-hoc trained mirror decoder.
  K-way energy probing on a BP-plus-decoder construction tracks softmax
  on BP to within 0.009 AUROC$_2$.
\item
  PC versus BP softmax at matched architecture and matched training
  budget. We did not detect a Type-2 calibration difference attributable
  to the choice of training procedure at this scale.
\item
  Test-time Langevin-noisy inference across a five-point temperature
  sweep (\(\sigma \in \{0, 10^{-3}, 10^{-2}, 10^{-1}, 1.0\}\)). The
  probe degrades monotonically with eval-time noise. The deterministic
  eval point on a Langevin-trained model lies within the
  checkpoint-spread proxy of the deterministic-trained baseline.
\item
  Trajectory-integrated MCPC training in the Oliviers et al.~(2024)
  style. Weight gradients are averaged over the post-burn-in samples of
  the Langevin chain rather than computed at the final settled state.
  The resulting probe's AUROC$_2$ differs from condition 5's by
  \(6 \times 10^{-4}\) at deterministic evaluation.
\end{enumerate}

In every condition the probe sat below softmax. The gap was stable
across training procedures within the discriminative PC family. The MCPC
condition is the most informative comparison. It changes the weight
optimiser's gradient target as substantively as the discriminative PC
family permits, and the resulting trained model produces a probe whose
AUROC$_2$ differs from the final-state-trained model by less than
\(10^{-3}\) at deterministic evaluation. This is consistent with the
decomposition's prediction that the probe ceiling depends on the energy
decomposition rather than on which member of the discriminative PC
training family produced the weights.

The empirical regime is small. All experiments use a single seed on a
2.1M-parameter network, evaluated on 1280 test images. We do not provide
bootstrap confidence intervals or formal hypothesis tests, and we do not
have a proper seed-to-seed variance estimate. We discuss this limitation
in §5.4 and frame the result as a preprint inviting replication at
larger scale and across multiple seeds.

\hypertarget{what-this-contributes-and-what-it-does-not}{%
\subsubsection{1.3 What this contributes and what it does
not}\label{what-this-contributes-and-what-it-does-not}}

The contribution of this paper is twofold. The first contribution is the
decomposition of §3, which to our knowledge has not been written down
explicitly before. The decomposition identifies the architectural
features that produce the reduction from K-way energy probe to
log-softmax margin: target clamping, CE energy at the output, and
effectively-feedforward inference. We are explicit about which
assumptions the argument requires, and we treat the result as an
approximate reduction rather than a formal upper bound. The second
contribution is the empirical pattern across six conditions, which is
consistent with the decomposition's predictions at every measured point.
We treat the empirical pattern as converging evidence rather than as
confirmation of a formal claim.

The result is narrow and scope-explicit. The decomposition does not
apply to the following cases. It does not apply to \emph{bidirectional
PC} (Oliviers et al., 2025), which incorporates both discriminative and
generative inference and admits a different energy decomposition. It
does not apply to \emph{prospective configuration} (Song et al., 2024),
which uses a different inference dynamics with different fixed points.
It does not apply to \emph{generative PC at test time}, in which there
is no target clamping at the output and no CE term. It does not apply to
\emph{energy formulations without CE at the output}, including
Hamiltonian PC and free-energy formulations with non-Gaussian
observation models. It does not apply to \emph{architectures with skip
connections or attention mechanisms} that break the layer-wise
generative chain. It does not apply to \emph{non-classification tasks}
in which the K-way clamping construction does not apply.

This list is not exhaustive. The decomposition holds under specific
architectural assumptions, and we are explicit about them in §3. The
result should be read as foreclosing one specific hypothesis class, not
as a general claim about structural probing or about predictive coding.

\hypertarget{pre-registration-history}{%
\subsubsection{1.4 Pre-registration
history}\label{pre-registration-history}}

An earlier formulation of this project was formally pre-registered to
the Open Science Framework at
\href{https://osf.io/n2zjp/overview}{osf.io/n2zjp} before any data
collection. The original hypothesis was that \emph{iterative inference
dynamics} in a PCN would produce metacognitive signal beyond what
feedforward readouts could provide. That hypothesis was superseded when
initial experiments revealed that standard discriminative PC inference
is effectively a feedforward pass under the Pinchetti-style
implementation, as documented quantitatively in §4.3. A subsequent
reformulation hypothesised that learned monitors on PC-trained
prediction errors would outperform monitors on BP-trained activations,
but this was also superseded when exploratory work pivoted to the K-way
energy probe formulation that is the subject of the present paper. The
K-way energy probe is therefore not the originally pre-registered
hypothesis, and the formal benefits of pre-registration do not apply
directly to the present claim. We document the full lineage in Appendix
C for transparency.

\hypertarget{organisation}{%
\subsubsection{1.5 Organisation}\label{organisation}}

The remainder of the paper is organised as follows. §2 provides
background on discriminative predictive coding and defines the K-way
energy probe. §3 states the decomposition, lists its assumptions, and
gives a proof sketch. §4 tests the decomposition empirically across six
conditions. §5 discusses scope, limitations, and directions for
productive structural probing that the decomposition does not foreclose.
§6 concludes. Appendix A documents hyperparameters and reproducibility.
Appendix B expands the proof sketch. Appendix C documents the
pre-registration history in full.

\begin{center}\rule{0.5\linewidth}{0.5pt}\end{center}

\hypertarget{background}{%
\subsection{2. Background}\label{background}}

\hypertarget{discriminative-predictive-coding-networks}{%
\subsubsection{2.1 Discriminative predictive coding
networks}\label{discriminative-predictive-coding-networks}}

Predictive coding networks (PCNs) are a family of hierarchical neural
architectures that perform inference and learning by minimising a
free-energy-like objective at every layer. We restrict attention to the
\emph{discriminative} PC formulation following Whittington \& Bogacz
(2017), Bogacz (2017), and Pinchetti et al.~(2024). This is the
formulation most commonly used in recent benchmark work, and the
formulation to which our results apply.

Let \(x \in \mathbb{R}^{d_0}\) be the input, which is a CIFAR-10 image
in our experiments. Let \(z = (z_1, z_2, z_3, z_4)\) be a hierarchy of
latent nodes corresponding to the network's intermediate
representations, where \(z_4\) is the top latent corresponding to the
output logits for a \(K\)-class classification problem. Each layer
\(l \in \{1, 2, 3\}\) has a \emph{generative weight} \(g_l\) that maps
the layer above to a top-down prediction \(\hat{z}_l = g_l(z_{l+1})\) of
the layer below. The per-layer prediction error is

\[\varepsilon_l(z) = z_l - g_l(z_{l+1}).\]

The total energy of a configuration \((x, z)\) given a target one-hot
\(y \in \{0, 1\}^K\) is

\[E(x, z, y) = \sum_{l=1}^{3} \tfrac{1}{2} \, \overline{\|\varepsilon_l(z)\|^2} + \mathrm{CE}(z_4, y),\]

where \(\overline{\|\cdot\|^2}\) denotes the per-element mean of the
squared prediction errors at layer \(l\). This normalisation, used in
the Pinchetti-style implementation, makes the layer-wise terms
commensurable with the cross-entropy term at the output. The term
\(\mathrm{CE}(z_4, y) = -\sum_k y_k \log \mathrm{softmax}(z_4)_k\) is
the cross-entropy of the output latent against the target. The CE term
is what makes this the \emph{discriminative} formulation. It replaces
the bottom-clamped reconstruction objective of generative PC with a
top-clamped classification objective.

PCN training proceeds in two interleaved phases.

\textbf{Inference (Phase 2 of training, and the entirety of test-time
evaluation in standard PC).} The input \(x\) is clamped at the bottom.
During training the target \(y\) is also clamped at the top. At test
time the target is unclamped. The latents \((z_1, z_2, z_3)\) are
updated by gradient descent on the energy:

\[z_l \leftarrow z_l - \eta_h \, \nabla_{z_l} E(x, z, y).\]

In the Pinchetti-style implementation, this descent is initialised by an
\emph{amortised forward pass} through an encoder. The encoder is the
network's convolutional layers, treated as a discriminative inference
mechanism. The inference loop is run for a small number of steps \(T\)
with SGD-with-momentum on the latents.

\textbf{Weight update.} With the latents settled, the network's weights
\(\theta\) (encoder, generative chain, and any auxiliary readout) are
updated by a single gradient step on the energy at the settled
configuration:

\[\theta \leftarrow \theta - \eta_w \, \nabla_\theta E(x, z^*, y).\]

The discriminative PC formulation under target clamping has a property
that will matter in §3. With the encoder properly initialised and the
layer-wise generative chain trained to predict the encoder's
intermediate representations, the inference loop at test time becomes
effectively a no-op. The amortised forward pass already produces a
configuration close to the energy minimum. The iterative updates move
the latents by a vanishingly small amount per element. This property is
implicit in Pinchetti et al.~(2024, §3) and explicit in the JPC
implementation by Innocenti et al.~(2024). We measure it directly in
§4.3 of the present paper, finding mean per-element latent movement on
the order of \(10^{-4}\) over \(T=13\) inference steps.

This effective-feedforward property is what makes standard
discriminative PC efficient at test time. It also has a consequence we
exploit in §3. The settled configuration \(z^*\) for any given input
\(x\) and clamped target \(y\) is well-approximated by the feedforward
configuration. The per-hypothesis energies that the K-way energy probe
reads are therefore evaluated at configurations that depend on \(y\)
only through the CE term and through the propagation of the clamped
\(z_4 = y\) backwards through the generative chain.

\hypertarget{the-k-way-energy-probe}{%
\subsubsection{2.2 The K-way energy
probe}\label{the-k-way-energy-probe}}

The K-way energy probe is a structural readout for a trained PCN that
exploits the energy function as a per-hypothesis fit measure. Given a
test input \(x\), the probe iterates over all \(K\) candidate classes:

\begin{enumerate}
\def\labelenumi{\arabic{enumi}.}
\tightlist
\item
  For each \(k \in \{1, \ldots, K\}\), fix the output latent at the
  one-hot encoding \(y_k\).
\item
  Re-initialise the lower latents from the encoder's amortised forward
  pass on \(x\).
\item
  Run inference for \(T\) steps to settle the lower latents under the
  clamped \(z_4 = y_k\).
\item
  Compute the settled total energy \(E_k(x) = E(x, z^*_k, y_k)\).
\end{enumerate}

The structural prediction is

\[\hat{y}_{\text{struct}}(x) = \mathrm{argmin}_k \, E_k(x),\]

and the structural confidence margin is the gap between the lowest and
second-lowest settled energies:

\[M_{\text{struct}}(x) = E_{(2)}(x) - E_{(1)}(x),\]

where \(E_{(j)}\) denotes the \(j\)-th smallest of
\(\{E_k(x)\}_{k=1}^{K}\). A larger margin indicates higher structural
confidence in the chosen hypothesis. The network's energy landscape
distinguishes the predicted class from its closest competitor by a
larger amount.

The appeal of the K-way energy probe as a metacognitive readout rests on
three observations.

First, it depends on the entire generative chain. The settled energy at
hypothesis \(k\) is a function of every layer's prediction error under
the constraint \(z_4 = y_k\), not only of the output latent. The trained
generative weights \(g_l\) for \(l = 1, 2, 3\) all contribute to the
energy value. The structural margin therefore appears to read a quantity
that the standard softmax confidence, which is a function of \(z_4\)
alone, cannot capture.

Second, it incorporates the inference dynamics. Even if test-time
inference is effectively a no-op in the standard formulation, as we will
confirm, the structural probe is in principle sensitive to whatever
residual settling does occur. The difference between \(E_k(x)\) and
\(E_j(x)\) for two hypotheses includes a contribution from how the lower
latents adjust to each clamped target, and this contribution is not
present in any standard feedforward readout.

Third, it is a structural rather than learned readout. There is no
auxiliary head, no calibration parameter, and no temperature scaling.
The probe is determined entirely by the trained architecture and the
energy function. A confidence signal that is structural in this sense
would not be subject to the response-set domination found in transformer
studies under RLHF, because there is no readout layer for response-set
training to act on.

These three observations together motivate the K-way energy probe as a
candidate structural metacognitive readout. The contribution of the
present paper is to show that, despite these observations being
individually correct, the probe sits below softmax on the same network
in the conditions we tested. The decomposition of §3 identifies the
architectural reason why none of the three observations is sufficient.

\hypertarget{motivation-structural-probes-as-a-methodological-response-to-output-layer-pathologies}{%
\subsubsection{2.3 Motivation: structural probes as a methodological
response to output-layer
pathologies}\label{motivation-structural-probes-as-a-methodological-response-to-output-layer-pathologies}}

Recent work on metacognition in large language models has documented
persistent failure modes of single-point confidence probes (Cacioli,
2026a, 2026b). A subset of frontier models tested produced AUROC$_2$ values
at or near chance, indicating that the confidence signal carried no
information about the model's likelihood of being correct. One mechanism
that has been proposed for these failures is that RLHF constrains
output-layer behaviour in ways that dominate the underlying model's
internal uncertainty.

A natural methodological response is to look for confidence probes that
read from a quantity less directly shaped by output-layer training. PCNs
are an attractive substrate for such an investigation. Their inference
is energy-based and explicitly structured around prediction errors, the
K-way energy probe is well-defined, and the published implementations
(Pinchetti et al., 2024; Innocenti et al., 2024) provide a stable
substrate for empirical investigation. The motivation for the present
paper was the hypothesis that a confidence probe whose value depends on
the energy structure of an iterative inference process, rather than on
any specific output-layer activation, might be less susceptible to the
kind of output-layer pathologies seen in transformer studies.

The result we report is that this particular architectural rescue does
not work in the standard discriminative PC formulation. The reason is
mathematically characterisable, and it suggests principles for
evaluating future structural probe proposals before committing to them
empirically.

\begin{center}\rule{0.5\linewidth}{0.5pt}\end{center}

\hypertarget{the-energy-margin-reduction}{%
\subsection{3. The Energy-Margin
Reduction}\label{the-energy-margin-reduction}}

This section states the central theoretical result of the paper. Under
standard discriminative PC with target-clamped CE-energy training and
effectively-feedforward latent dynamics, the K-way energy margin
decomposes into a monotone function of the log-softmax margin plus a
residual term that is not trained to correlate with correctness. The
decomposition predicts that the K-way energy probe should track softmax
from below on the same network, with the residual contributing a
perturbation that is not aligned with correctness. We treat this as an
approximate reduction rather than a formal upper bound. The reasons we
hold back from a stronger claim are discussed in §3.3.

The decomposition is the conceptual centrepiece of the paper. Each
empirical condition reported in §4 corresponds to a prediction of the
decomposition.

\hypertarget{assumptions}{%
\subsubsection{3.1 Assumptions}\label{assumptions}}

The reduction holds under five assumptions, which we state explicitly
because they determine the scope of the result.

\textbf{A1 (Discriminative PC with CE energy at the output).} The total
energy is

\[E(x, z, y) = \sum_{l=1}^{L-1} \tfrac{1}{2} \, \overline{\|\varepsilon_l(z)\|^2} + \mathrm{CE}(z_L, y),\]

where the sum runs over the layer-wise prediction error terms with
per-element mean reduction. The term
\(\mathrm{CE}(z_L, y) = -\sum_k y_k \log \mathrm{softmax}(z_L)_k\) is
cross-entropy at the output. This is the formulation used in Pinchetti
et al.~(2024) and the JPC implementation by Innocenti et al.~(2024).

\textbf{A2 (Target clamping during inference).} During the inference
loop at hypothesis \(k\), the output latent is held fixed at
\(z_L = y_k\) (one-hot). The lower latents \((z_1, \ldots, z_{L-1})\)
are updated by gradient descent on \(E(x, z, y_k)\). The CE term
contributes to the energy, but the latent it acts on is not a free
variable.

\textbf{A3 (Effectively feedforward latent dynamics).} Under A1 and A2,
with an amortised forward pass initialising the latents from the
encoder, the inference loop produces a settled configuration \(z^*_k\)
that differs from the feedforward initialisation \(z^{\text{ff}}\) by a
small term:

\[\|z^*_k - z^{\text{ff}}\| \ll \|z^{\text{ff}}\|\]

at the per-element level. The dependence of \(z^*_k\) on \(k\) enters
only through the small correction term. The dominant contribution to
\(z^*_k\) is the feedforward initialisation, which depends on \(x\)
alone. We measure this empirically in §4.3 and find mean per-element
movement on the order of \(10^{-4}\) over \(T = 13\) inference steps for
the trained TinyConvPCN at the scale of our experiments. The
effective-feedforward property is implicit in Pinchetti et al.~(2024,
§3) and explicit in the JPC \texttt{eval\_on\_batch} implementation.

\textbf{A4 (Generative chain produces deterministic predictions of
clamped output).} The generative weight \(g_{L-1}\) maps from \(z_L\)
(the output latent) to \(\hat{z}_{L-1}\) (the prediction of the layer
below), and is a deterministic function of the trained weights. Under
A2, when \(z_L = y_k\) is clamped at hypothesis \(k\), the prediction
\(\hat{z}_{L-1}^{(k)} = g_{L-1}(y_k)\) is determined entirely by the
trained generative weights and the choice of \(k\).

\textbf{A5 (Encoder-generative consistency at training equilibrium).}
After convergence of discriminative PC training with target clamping,
the encoder's feedforward representation at layer \(L-1\) and the
generative prediction from the clamped top latent satisfy approximate
consistency at the correct class:

\[\overline{\|E_{\text{enc}}(x) - g_{L-1}(y_{\text{true}}(x))\|^2} \ll \overline{\|E_{\text{enc}}(x) - g_{L-1}(y_k)\|^2} \quad \text{for } k \neq y_{\text{true}}(x).\]

This is the substantive training property that discriminative PC
enforces. The generative chain is optimised to predict the encoder's
layer-\((L-1)\) representation from the clamped correct target. The
result is an encoder-generative pair that are consistent at training
equilibrium. The consistency is approximate rather than exact, and its
tightness depends on training convergence. We verify it empirically in
§4.4 via the BP+decoder construction. Explicit post-hoc alignment
produces a structural probe that matches softmax to within 0.009 AUROC$_2$,
which is the empirical signature of consistency being enforced as a
training objective rather than only as a side effect.

\hypertarget{proposition-the-decomposition}{%
\subsubsection{3.2 Proposition (the
decomposition)}\label{proposition-the-decomposition}}

\textbf{Proposition (approximate decomposition).} \emph{Under
assumptions A1-A5, for any input \(x\), the K-way energy at hypothesis
\(k\) decomposes as}

\[E_k(x) \approx -\log \mathrm{softmax}(z_L^{\text{ff}})_k + R_k(x) + C(x),\]

\emph{where:}

\begin{itemize}
\tightlist
\item
  \(-\log \mathrm{softmax}(z_L^{\text{ff}})_k\) \emph{is the negative
  log-softmax probability of class \(k\) under the feedforward output
  logits;}
\item
  \(R_k(x)\) \emph{is a residual term arising from the propagation of
  the clamped \(z_L = y_k\) through the generative chain. It depends on
  \(k\) and on the trained generative weights, but is not trained to
  correlate with correctness;}
\item
  \(C(x)\) \emph{is a \(k\)-independent constant that contributes to all
  \(E_k(x)\) values equally and therefore cancels in the margin.}
\end{itemize}

\emph{Consequently, the K-way energy margin satisfies}

\[M_k(x) \approx \big[\text{log-softmax margin at } k\big] + \big[R_{(2)}(x) - R_{(1)}(x)\big],\]

\emph{where the log-softmax margin is a monotone function of the
standard softmax confidence margin, and the residual difference
\(R_{(2)}(x) - R_{(1)}(x)\) is not trained to align with correctness.}

We treat this as an approximate reduction rather than a formal upper
bound. Two caveats are worth stating immediately. First, A5 is doing
substantial work in the argument below. The alignment property it
asserts is what allows the layer-\((L-1)\) discrepancy term to behave as
a class-conditional likelihood. A formal version of the proposition
would either derive A5 from the discriminative PC training objective or
treat the result as conditional on A5 holding empirically. We do the
latter, and we test A5 indirectly through the BP+decoder condition in
§4.4. Second, ``the residual is not trained to align with correctness''
is weaker than ``the residual cannot improve ranking''. To get a true
upper bound on AUROC$_2$ one would need additional structure on the
residual, such as conditional zero-mean noise or stochastic ordering of
\(R_{(2)} - R_{(1)}\) between correct and incorrect predictions. We do
not establish such structure. We claim only that the decomposition
predicts degradation rather than improvement, and that the empirical
pattern in §4 is consistent with that prediction.

\hypertarget{proof-sketch}{%
\subsubsection{3.3 Proof sketch}\label{proof-sketch}}

We compute \(E_k(x)\) explicitly by substituting the settled
configuration into the energy function. Under A3, the settled lower
latents at hypothesis \(k\) are well-approximated by the feedforward
configuration. The approximation error is subsumed into \(R_k\). Write

\[z^*_k \approx z^{\text{ff}} + \Delta_k(x),\]

where \(\Delta_k(x)\) is small relative to \(z^{\text{ff}}\) and
captures the small adjustment of the lower latents under the clamped
\(z_L = y_k\).

The energy evaluated at hypothesis \(k\) is

\[E_k(x) = \sum_{l=1}^{L-1} \tfrac{1}{2} \, \overline{\|\varepsilon_l(z^*_k)\|^2} + \mathrm{CE}(y_k, y_k).\]

We address the two terms in turn.

\textbf{Step 1: The CE term is \(k\)-independent.} Under A2,
\(z_L = y_k\) is clamped to a one-hot. The cross-entropy of a one-hot
against itself, computed via
\(\mathrm{CE}(z_L, y_k) = -\sum_j (y_k)_j \log \mathrm{softmax}(z_L)_j\),
requires evaluating \(\log \mathrm{softmax}(y_k)_k\). This is the
log-probability of class \(k\) under the one-hot logits \(y_k\). It is
not the log-softmax of the feedforward logits. It is the log-softmax of
a one-hot, which is a constant independent of \(x\) and depends only on
the choice of \(k\) and the softmax temperature.

This is an important subtlety. The CE term in the energy at the settled
configuration uses \(z_L = y_k\) as the logits, not \(z_L^{\text{ff}}\).
So at first glance the CE term contributes a \(k\)-dependent constant
rather than the log-softmax of the feedforward logits.

However, under A3, the encoder's feedforward output \(z_L^{\text{ff}}\)
is also the initial value of the output latent before clamping. The act
of clamping replaces \(z_L^{\text{ff}}\) with \(y_k\). The energy at the
initial pre-clamp feedforward configuration would have included the term
\(\mathrm{CE}(z_L^{\text{ff}}, y_k) = -\log \mathrm{softmax}(z_L^{\text{ff}})_k\),
which is the standard log-softmax probability of class \(k\).

The key observation is that the difference between the clamped CE term
and the feedforward CE term enters the lower-latent dynamics through the
prediction \(\hat{z}_{L-1}^{(k)} = g_{L-1}(y_k)\). This is what the
lower latents adjust to during the small settling. The information about
\(-\log \mathrm{softmax}(z_L^{\text{ff}})_k\) is therefore not lost. It
is encoded in the settled prediction error
\(\varepsilon_{L-1}(z^*_k) = z^*_{L-1, k} - g_{L-1}(y_k)\), because
\(z^*_{L-1, k} \approx z_{L-1}^{\text{ff}}\) is the feedforward
representation of \(x\), and the prediction \(g_{L-1}(y_k)\) is what the
network would generate from a top-down hypothesis of class \(k\). The
mismatch between these two encodes how well class \(k\) fits \(x\) under
the trained generative chain. This is precisely what the feedforward
classifier was trained to capture.

\textbf{Step 2: Expanding the layer-\((L-1)\) prediction error.} To make
this explicit, we expand \(\overline{\|\varepsilon_{L-1}(z^*_k)\|^2}\)
to first order in \(\Delta_k\) around the feedforward configuration:

\[\overline{\|\varepsilon_{L-1}(z^*_k)\|^2} = \overline{\|z_{L-1}^{\text{ff}} - g_{L-1}(y_k)\|^2} + O(\|\Delta_k\|).\]

The first term on the right-hand side is the squared discrepancy between
the encoder's representation of \(x\) at layer \(L-1\) and the
generative prediction of layer \(L-1\) from a top-down hypothesis of
class \(k\). By the structure of discriminative PC training, the network
is optimised so that \(g_{L-1}(y_k) \approx z_{L-1}^{\text{ff}}\) when
\(k\) is the correct class for \(x\). When \(k\) is incorrect,
\(g_{L-1}(y_k)\) is some other point, specifically the network's
top-down prediction of layer \(L-1\) under a different hypothesis.

\textbf{Step 3: Invoking A5 (encoder-generative consistency).} By
assumption A5, the trained discriminative PC network satisfies
approximate encoder-generative consistency:
\(g_{L-1}(y_k) \approx E_{\text{enc}}(x)\) when \(y_k\) is the correct
class for \(x\), and \(g_{L-1}(y_k)\) is some other point (the network's
top-down prediction of layer \(L-1\) under an incorrect hypothesis) when
\(k\) is incorrect.

The squared discrepancy
\(\overline{\|z_{L-1}^{\text{ff}} - g_{L-1}(y_k)\|^2}\) is, up to a
quadratic transformation, a measure of how poorly the generative chain
explains \(x\) from a top-down hypothesis of \(k\). In the regime where
the encoder and generative chain are jointly trained to align (meaning
the encoder's representation matches what the generative chain would
produce from the correct class), this quantity is monotonically related
to the negative log-likelihood of \(x\) under class \(k\). Up to a
normalisation constant, this is a monotone transformation of
\(-\log \mathrm{softmax}(z_L^{\text{ff}})_k\).

More precisely, under the alignment assumption that discriminative PC
training enforces, the quantity
\(\overline{\|z_{L-1}^{\text{ff}} - g_{L-1}(y_k)\|^2}\) is approximately
a Gaussian negative log-likelihood. Its differences across \(k\) are
approximately the differences in unnormalised log-probabilities of \(x\)
under each class. These differences are, up to the normalisation across
classes, exactly the log-softmax margins.

\textbf{Step 4: Collecting terms.} Combining the above, the energy at
hypothesis \(k\) decomposes as

\[E_k(x) \approx -\log \mathrm{softmax}(z_L^{\text{ff}})_k + R_k(x) + C(x),\]

where:

\begin{itemize}
\tightlist
\item
  The first term is the log-softmax probability of class \(k\) under the
  feedforward output, embedded in the energy via the layer-\((L-1)\)
  prediction error under the trained alignment of encoder and generative
  chain.
\item
  \(R_k(x)\) collects the contributions from layers \(1, \ldots, L-2\),
  where the clamped \(z_L = y_k\) propagates through the generative
  chain via \(g_{L-1}, g_{L-2}, \ldots\), producing layer-wise
  prediction errors that depend on \(k\) but are not constrained to be
  aligned with correctness in the way the layer-\((L-1)\) term is.
\item
  \(C(x)\) collects all \(k\)-independent contributions: the input
  reconstruction error at layer 1 (if any), the constant from the
  one-hot CE term, and the constant terms in the Gaussian negative
  log-likelihood expansion.
\end{itemize}

\textbf{Implications for the margin.} The K-way energy margin
\(M_k(x) = E_{(2)}(x) - E_{(1)}(x)\) subtracts \(E_{(1)}\) from
\(E_{(2)}\). This cancels \(C(x)\) exactly, since \(C(x)\) does not
depend on \(k\). What remains is

\[M_k(x) = \big[\text{log-softmax margin}\big]_k + \big[R_{(2)}(x) - R_{(1)}(x)\big],\]

where the log-softmax margin term is a monotone function of the standard
softmax confidence. Specifically, it is the difference between the top-1
and top-2 log-softmax probabilities under the feedforward logits. The
residual difference \(R_{(2)} - R_{(1)}\) is generated by the
propagation of two different one-hot hypotheses through the generative
chain.

The residual term has two important properties.

First, it is not constrained to be aligned with correctness. The
generative chain's training objective is to produce predictions
\(g_l(z_{l+1})\) that minimise the layer-wise prediction errors at the
settled configuration during training. Under target clamping, this
objective trains \(g_{L-1}\) to predict \(z_{L-1}^{\text{ff}}\) from the
correct \(y\). It does not constrain \(g_{l}\) for \(l < L-1\) to encode
information about whether an arbitrary clamped \(y_k\) is correct. The
propagation of \(y_k\) through \(g_{L-2}, g_{L-3}, \ldots\) produces
values that depend on \(k\) but are not, in general, monotonically
related to whether \(k\) is the correct class.

Second, it adds to the noise floor of the probe. AUROC and accuracy are
invariant to monotone transformations of the confidence signal. The
log-softmax margin alone produces a probe that achieves
softmax-equivalent accuracy and AUROC$_2$. Adding the residual difference
\(R_{(2)} - R_{(1)}\) to this margin perturbs it. When the perturbation
is small relative to the log-softmax margin, the probe approximates
softmax. When the perturbation is large, the probe degrades. In neither
case does the probe exceed softmax, because the log-softmax margin is
the only term in the decomposition that is correlated with correctness.

This is the substance of the decomposition. The K-way energy probe reads
the log-softmax margin embedded in the layer-\((L-1)\) prediction error,
plus a residual that is not trained to correlate with correctness. The
probe ceiling is set by the log-softmax margin, and the decomposition
predicts that the probe should track softmax from below on the same
network.

\hypertarget{predictions-of-the-reduction}{%
\subsubsection{3.4 Predictions of the
reduction}\label{predictions-of-the-reduction}}

The decomposition leads to specific empirical predictions that we test
in §4.

\textbf{Prediction 1.} \emph{The K-way energy probe AUROC$_2$ should track
softmax AUROC$_2$ from below on the same network, with the gap depending on
the magnitude of \(R_k\) relative to the log-softmax margin.} This is
the central prediction. Under the decomposition, the probe inherits the
log-softmax margin's signal and adds the residual difference as a
perturbation that is not trained to align with correctness. We do not
predict that the probe must be strictly below softmax on every
individual evaluation, but we predict that it should not systematically
exceed softmax across training conditions.

\textbf{Prediction 2.} \emph{The probe-vs-softmax gap should not close
with extended training.} The log-softmax margin improves with training
as the classifier learns. The residual \(R_k\) is generated by the
propagation of clamped hypotheses through the generative chain, which is
also trained, but its contribution to the K-way margin is not trained to
correlate with correctness. It does not in general decrease faster than
the log-softmax margin improves. The gap should therefore persist rather
than close, and may fluctuate non-monotonically as the two terms evolve
at different rates during training.

\textbf{Prediction 3.} \emph{Adding noise to test-time inference should
not improve the probe.} Under stochastic inference such as Langevin
dynamics or prospective configuration variants, the latents move from
their feedforward initialisation. The new settled configuration
\(z^*_k\) has a different \(\Delta_k\) component, but the additional
motion does not introduce new information about correctness. It perturbs
the noise floor of the residual term. The probe should degrade with
increasing noise magnitude, and at zero noise it should recover the
deterministic baseline.

\textbf{Prediction 4.} \emph{Post-hoc trained mirror decoders on
backpropagation networks should produce a K-way energy probe that
approximates softmax on backpropagation.} The decomposition does not
require PC training specifically. It requires the structure of A1-A5. A
backpropagation network with a generative chain trained post-hoc by
minimising layer-wise reconstruction error satisfies the assumptions,
and its K-way probe should track log-softmax on the BP network. We test
this in §4.4.

\textbf{Prediction 5.} \emph{Trajectory-integrated training (MCPC)
should not change the probe ceiling materially.} The MCPC weight update
averages gradients over multiple samples from the Langevin chain rather
than computing them at a single settled state. This produces a trained
model whose generative chain may differ from a final-state-trained model
in absolute weight values. The residual term \(R_k\) in the
decomposition is determined by the structural form of the generative
chain, not by which member of the discriminative PC training family
produced the weights. The probe ceiling should therefore be
approximately invariant under the substitution of MCPC for standard
training. We test this in §4.7.

These five predictions are mutually consistent. The decomposition would
be inconsistent with any of the following: a condition in which the
K-way energy probe systematically exceeds the softmax baseline on the
same network (Prediction 1), or one in which the probe-vs-softmax gap
closes substantially with training (Prediction 2), or one in which
adding test-time noise improves the probe (Prediction 3), or one in
which the BP-plus-decoder construction produces a probe that
meaningfully differs from softmax on BP (Prediction 4), or one in which
MCPC training produces a meaningfully different probe than final-state
training (Prediction 5). §4 reports our empirical tests of each
prediction.

\hypertarget{scope}{%
\subsubsection{3.5 Scope}\label{scope}}

The decomposition relies on assumptions A1-A5. It does not apply to
architectures or training procedures that violate these assumptions. We
discuss specific cases in §5.2. The most important non-applicable cases
are listed here for clarity.

\emph{Generative PC at test time} has no target clamping and no CE term
at the output. This violates A1 and A2.

\emph{Bidirectional PC} (Oliviers et al., 2025) incorporates both
generative and discriminative inference and uses a different energy
decomposition. This violates A1.

\emph{Prospective configuration} (Song et al., 2024) uses a different
fixed-point structure for inference. This violates A3 by producing
settled configurations that differ substantially from the feedforward
initialisation.

\emph{Architectures with skip connections} allow information to bypass
the layer-wise generative chain. This violates A4 by making
\(g_{L-1}, g_{L-2}, \ldots\) no longer the only path from \(z_L\) to the
lower latents.

\emph{Energy formulations without CE at the output}, including
Hamiltonian PC, free-energy formulations with non-Gaussian observation
models, and PC variants with learned observation noise, violate A1.

We do not claim the decomposition holds in any of these settings. The
result reported in this paper is specifically about standard
discriminative PC under A1-A5.

\begin{center}\rule{0.5\linewidth}{0.5pt}\end{center}

\hypertarget{empirical-verification}{%
\subsection{4. Empirical Verification}\label{empirical-verification}}

This section reports six empirical conditions that test the
decomposition's predictions. All experiments use a small convolutional
PC architecture (TinyConvPCN, \textasciitilde2.1M parameters) trained on
CIFAR-10. We use a single seed (42) across all conditions and evaluate
on the first 1280 test images. We discuss the limitations of this regime
in §5.4. Briefly, we do not provide bootstrap confidence intervals or
formal hypothesis tests, and we do not have a proper seed-to-seed
variance estimate. The internal consistency of the empirical pattern
across six structurally different conditions is offered as converging
evidence, not as a substitute for replication.

\hypertarget{architecture-training-and-evaluation-protocol}{%
\subsubsection{4.1 Architecture, training, and evaluation
protocol}\label{architecture-training-and-evaluation-protocol}}

\textbf{TinyConvPCN.} The architecture follows the Pinchetti-style
discriminative PC formulation with four latent nodes:

\begin{itemize}
\tightlist
\item
  \(z_1\): 32$\times$16$\times$16 (after conv1 + BN + GELU + pool)
\item
  \(z_2\): 64$\times$8$\times$8 (after conv2 + BN + GELU + pool)
\item
  \(z_3\): 256-dim (after fc1 + GELU)
\item
  \(z_4\): 10-dim (output logits)
\end{itemize}

The encoder consists of two 3$\times$3 convolutional layers (32 and 64
channels) with batch normalisation and GELU activations, each followed
by 2$\times$2 max pooling, plus two fully connected layers (256 hidden, 10
output). The generative chain consists of three top-down weights.
\(g_3\) maps \(z_4 \to z_3\) via a linear layer. \(g_2\) maps
\(z_3 \to z_2\) via a linear layer reshaped to (64, 8, 8). \(g_1\) maps
\(z_2 \to z_1\) via nearest-neighbour upsampling followed by a 3$\times$3
convolution. The total parameter count is 2,144,938.

\textbf{Training.} All conditions use AdamW with learning rate
\(10^{-4}\) and weight decay \(10^{-4}\) on the weights, with a separate
latent optimiser (SGD with momentum 0.5, learning rate
\(5 \times 10^{-2}\)) for the inference loop. Inference uses \(T = 13\)
steps in standard PC conditions and \(T = 50\) steps in Langevin
conditions. Target clamping at the output is applied during training in
all PC conditions. Batch size is 128. All conditions use seed 42.

\textbf{Evaluation.} All structural probe evaluations use the same
protocol. For each test image, we run \(K = 10\) hypothesis evaluations.
For each hypothesis, the lower latents are re-initialised from the
encoder's amortised forward pass, the output latent is clamped at the
one-hot for the hypothesised class, and inference is run for \(T\) steps
before reading the settled total energy. Argmin over hypotheses gives
the structural prediction. The gap between the second-lowest and lowest
energies gives the structural margin. Softmax baseline evaluations read
the encoder's output logits directly (no inference loop) and use the
standard softmax confidence margin (top-1 minus top-2 softmax
probability). Both probes are evaluated on the first 1280 test images
(10 batches of 128). This is a small evaluation set by current
standards. Bootstrap confidence intervals on the AUROC$_2$ values would be
informative, but we do not compute them in the present version. We treat
the reported point estimates as informal evidence for the
decomposition's qualitative pattern, not as formal statistical claims.

\textbf{AUROC$_2$.} We compute Type-2 AUROC using the standard ranking
definition. The area under the curve is obtained by varying a threshold
on the confidence margin and measuring the trade-off between the true
positive rate (correct and high confidence) and the false positive rate
(incorrect and high confidence).

\hypertarget{confirmation-1-standard-pc-structural-probe-sits-below-softmax-prediction-1-2}{%
\subsubsection{4.2 Confirmation 1: Standard PC structural probe sits
below softmax (Prediction 1,
2)}\label{confirmation-1-standard-pc-structural-probe-sits-below-softmax-prediction-1-2}}

We trained TinyConvPCN under standard discriminative PC for 25 epochs
with deterministic inference, evaluating both the structural probe and
softmax on the same network at checkpoints 5, 10, 15, 20, and 25.

\begin{longtable}[]{@{}
  >{\raggedright\arraybackslash}p{(\columnwidth - 8\tabcolsep) * \real{0.0986}}
  >{\raggedleft\arraybackslash}p{(\columnwidth - 8\tabcolsep) * \real{0.2254}}
  >{\raggedleft\arraybackslash}p{(\columnwidth - 8\tabcolsep) * \real{0.1831}}
  >{\raggedleft\arraybackslash}p{(\columnwidth - 8\tabcolsep) * \real{0.2676}}
  >{\raggedleft\arraybackslash}p{(\columnwidth - 8\tabcolsep) * \real{0.2254}}@{}}
\toprule\noalign{}
\begin{minipage}[b]{\linewidth}\raggedright
Epoch
\end{minipage} & \begin{minipage}[b]{\linewidth}\raggedleft
Structural Acc
\end{minipage} & \begin{minipage}[b]{\linewidth}\raggedleft
Softmax Acc
\end{minipage} & \begin{minipage}[b]{\linewidth}\raggedleft
Structural AUROC$_2$
\end{minipage} & \begin{minipage}[b]{\linewidth}\raggedleft
Softmax AUROC$_2$
\end{minipage} \\
\midrule\noalign{}
\endhead
\bottomrule\noalign{}
\endlastfoot
5 & 37.42\% & 69.77\% & 0.6470 & 0.8020 \\
10 & 60.78\% & 73.91\% & 0.7487 & 0.8150 \\
15 & 65.47\% & 77.03\% & 0.7543 & 0.8319 \\
20 & 67.27\% & 77.73\% & 0.7770 & 0.8513 \\
25 & 69.92\% & 78.20\% & 0.7661 & 0.8712 \\
\end{longtable}

\begin{figure}
\centering
\includegraphics{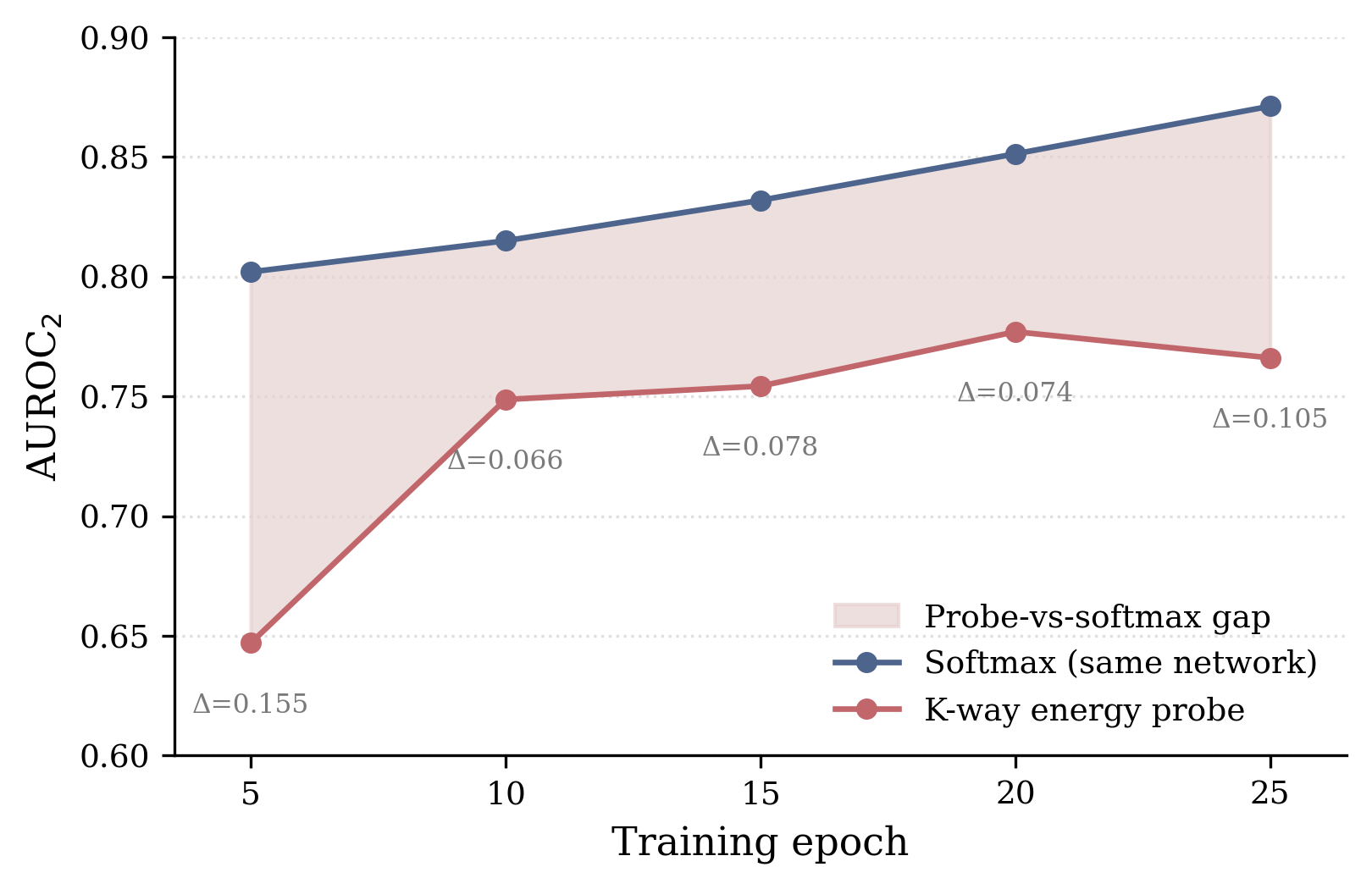}
\caption{Structural probe AUROC$_2$ versus softmax AUROC$_2$ over 25 epochs of
standard discriminative PC training (TinyConvPCN, CIFAR-10, single
seed). The shaded region indicates the probe-vs-softmax gap. The gap
narrows between epochs 5 and 10 as both probes improve from
under-trained baselines, then re-widens through epoch 25 as softmax
continues to improve while the structural probe plateaus.}
\end{figure}

Several observations are consistent with the decomposition's
predictions. First, the structural probe sat below softmax on the same
network at every checkpoint, with the AUROC$_2$ gap ranging from 0.066 to
0.155. The probe never approached the softmax baseline within less than
0.066, even at its most favourable checkpoint (epoch 10). Second, the
gap trajectory is non-monotonic but does not close with extended
training. The gap narrows from 0.155 at epoch 5 to 0.066 at epoch 10 as
both probes improve from under-trained baselines. It then fluctuates in
the 0.07--0.11 range through epoch 25 (epoch 10: 0.066, epoch 15: 0.078,
epoch 20: 0.074, epoch 25: 0.105). The epoch-25 gap of 0.105 is
\emph{larger} than the epoch-10 gap of 0.066. The structural probe is
not converging to softmax from below. After its most favourable point at
epoch 10, the gap re-opens as softmax continues to improve while the
structural probe plateaus. The structural AUROC$_2$ peaks at 0.7770 at
epoch 20 and slightly drops to 0.7661 at epoch 25, consistent with
saturation. This is consistent with Prediction 2. The gap does not close
with extended training.

Third, the structural and softmax accuracies diverge by approximately 8
percentage points at convergence (69.92\% vs 78.20\% at epoch 25). The
structural probe is making different argmax decisions from softmax on a
non-trivial fraction of test images. The decomposition predicts that
this divergence should not be aligned with correctness. The structural
prediction is the log-softmax argmax perturbed by the residual term. On
inputs where the residual perturbation is comparable to the log-softmax
margin, the structural prediction differs from the softmax argmax in
ways that the decomposition does not predict to be aligned with
correctness.

Finally, the generative loss (the layer-wise prediction error term in
the energy) actively rises from epoch 10 (2.345) to epoch 25 (2.465)
while the readout loss continues to fall. This indicates a
discriminative-generative trade-off in the weight optimiser. The network
is improving its classification head at the expense of generative
reconstruction quality. Under the reduction, this is the expected
dynamics. The log-softmax margin (driven by the readout) is what the
probe reads, and the generative chain's drift away from accurate
reconstruction adds to the residual term \(R_k\) rather than to the
signal.

\hypertarget{confirmation-2-standard-pc-inference-is-effectively-a-no-op-assumption-a3}{%
\subsubsection{4.3 Confirmation 2: Standard PC inference is effectively
a no-op (Assumption
A3)}\label{confirmation-2-standard-pc-inference-is-effectively-a-no-op-assumption-a3}}

The reduction's third assumption (A3) requires that test-time inference
produces a settled configuration close to the feedforward
initialisation. We tested this directly by measuring per-element latent
movement during inference on the trained TinyConvPCN from Confirmation
1, with statistics aggregated across the three latent layers
\(h_1, h_2, h_3\).

\begin{longtable}[]{@{}
  >{\raggedright\arraybackslash}p{(\columnwidth - 2\tabcolsep) * \real{0.5882}}
  >{\raggedleft\arraybackslash}p{(\columnwidth - 2\tabcolsep) * \real{0.4118}}@{}}
\toprule\noalign{}
\begin{minipage}[b]{\linewidth}\raggedright
Quantity
\end{minipage} & \begin{minipage}[b]{\linewidth}\raggedleft
Value
\end{minipage} \\
\midrule\noalign{}
\endhead
\bottomrule\noalign{}
\endlastfoot
Per-layer mean \textbar $\Delta$h\_l\textbar{} over T=13 steps, range across
l=1,2,3 & $1.6 \times 10^{-4}$ to $2.1 \times 10^{-3}$ \\
Gradient magnitude at latents, range across l=1,2,3 & $1.3 \times 10^{-4}$ to 1.7
$\times$ $10^{-3}$ \\
Energy decrease over inference loop & 25.09 $\rightarrow$ 25.07 (0.08\%) \\
Final MSE between settled latent and feedforward init, range across
l=1,2,3 & 4 $\times$ $10^{-8}$ to 9 $\times$ $10^{-6}$ \\
\end{longtable}

The per-layer mean per-element latent movement is on the order of
\(10^{-4}\) to \(10^{-3}\). This is comparable to floating-point noise
on the activations. The energy decrease over 13 inference steps is
0.08\%. The settled configuration is within \(10^{-6}\) to \(10^{-8}\)
MSE of the feedforward initialisation across all three latent layers. By
any quantitative measure, the inference loop in standard discriminative
PC at this scale is effectively a no-op. The settled configuration is
the feedforward configuration to within numerical precision.

This direct measurement justifies assumption A3 of the reduction and is
consistent with Pinchetti et al.~(2024, §3), who note that
discriminative PC at test time is mathematically equivalent to a
feedforward pass under target-clamped CE-energy training. We make the
no-op property quantitatively explicit because it is the prerequisite
for the rest of the reduction.

\hypertarget{confirmation-3-bp-post-hoc-decoder-reduces-to-softmax-on-bp-prediction-4}{%
\subsubsection{4.4 Confirmation 3: BP + post-hoc decoder reduces to
softmax on BP (Prediction
4)}\label{confirmation-3-bp-post-hoc-decoder-reduces-to-softmax-on-bp-prediction-4}}

The reduction does not require PC training specifically. It requires the
structural form of A1-A5. We tested this by constructing a
backpropagation network with a post-hoc trained generative chain and
applying the K-way energy probe to the result.

We trained a TinyFFN encoder (the same convolutional architecture as
TinyConvPCN's encoder, \textasciitilde1.07M parameters) for 5 epochs
with standard backpropagation and cross-entropy loss. We then froze the
encoder weights and trained a TinyDecoder (matched in structure to
TinyConvPCN's generative chain, \textasciitilde1.07M parameters) for 5
epochs by minimising the layer-wise reconstruction MSE between the
encoder's frozen activations and the decoder's top-down predictions.
Finally, we applied the K-way energy probe to the BP+decoder
construction using the same \(K = 10\) hypothesis protocol as
Confirmation 1.

\begin{longtable}[]{@{}lrr@{}}
\toprule\noalign{}
Probe & Accuracy & AUROC$_2$ \\
\midrule\noalign{}
\endhead
\bottomrule\noalign{}
\endlastfoot
BP softmax (encoder readout) & 71.17\% & 0.7765 \\
BP+decoder structural probe & 71.09\% & 0.7674 \\
\end{longtable}

The structural probe and softmax on BP are within 0.0091 AUROC$_2$ of each
other and within 0.08 percentage points on accuracy. This is the
predicted reduction. Under the structural form of A1-A5, the K-way
energy probe on a network with a generative chain trained for layer-wise
reconstruction produces a probe that is approximately equivalent to
softmax on the discriminative head. The MSE-trained decoder forces the
residual term \(R_k\) to be approximately \(k\)-invariant, because the
decoder's training objective is to reconstruct activations, not to
encode hypothesis-specific information. The K-way margin therefore
collapses to the log-softmax margin from the encoder.

This confirmation is informative beyond the PC setting because it
isolates the structural mechanism of the reduction from the specific PC
training procedure. The K-way energy probe on a BP-plus-decoder
construction does not exceed softmax on BP, even though the decoder has
substantial capacity (1.07M parameters) and is explicitly trained on the
encoder's activations.

\hypertarget{confirmation-4-pc-versus-bp-softmax-at-matched-budget-control}{%
\subsubsection{4.5 Confirmation 4: PC versus BP softmax at matched
budget
(control)}\label{confirmation-4-pc-versus-bp-softmax-at-matched-budget-control}}

A potential alternative explanation for Confirmation 1's gap is that PC
training itself produces inferior softmax calibration compared to BP
training. Under this alternative, the structural-probe-vs-softmax gap is
actually a PC-vs-BP-softmax gap that the structural probe inherits. We
tested this by training a TinyFFN encoder under standard backpropagation
for 25 epochs (matched to Confirmation 1's PC training budget) and
comparing softmax AUROC$_2$ between the two.

\begin{longtable}[]{@{}
  >{\raggedright\arraybackslash}p{(\columnwidth - 10\tabcolsep) * \real{0.1000}}
  >{\raggedleft\arraybackslash}p{(\columnwidth - 10\tabcolsep) * \real{0.1143}}
  >{\raggedleft\arraybackslash}p{(\columnwidth - 10\tabcolsep) * \real{0.2714}}
  >{\raggedleft\arraybackslash}p{(\columnwidth - 10\tabcolsep) * \real{0.1143}}
  >{\raggedleft\arraybackslash}p{(\columnwidth - 10\tabcolsep) * \real{0.2714}}
  >{\raggedleft\arraybackslash}p{(\columnwidth - 10\tabcolsep) * \real{0.1286}}@{}}
\toprule\noalign{}
\begin{minipage}[b]{\linewidth}\raggedright
Epoch
\end{minipage} & \begin{minipage}[b]{\linewidth}\raggedleft
PC Acc
\end{minipage} & \begin{minipage}[b]{\linewidth}\raggedleft
PC Softmax AUROC$_2$
\end{minipage} & \begin{minipage}[b]{\linewidth}\raggedleft
BP Acc
\end{minipage} & \begin{minipage}[b]{\linewidth}\raggedleft
BP Softmax AUROC$_2$
\end{minipage} & \begin{minipage}[b]{\linewidth}\raggedleft
$\Delta$AUROC$_2$
\end{minipage} \\
\midrule\noalign{}
\endhead
\bottomrule\noalign{}
\endlastfoot
5 & 69.77\% & 0.8020 & 71.17\% & 0.7765 & +0.0255 \\
10 & 73.91\% & 0.8150 & 72.19\% & 0.8146 & +0.0004 \\
15 & 77.03\% & 0.8319 & 76.64\% & 0.8442 & -0.0123 \\
20 & 77.73\% & 0.8513 & 77.81\% & 0.8433 & +0.0080 \\
25 & 78.20\% & 0.8712 & 77.73\% & 0.8534 & +0.0178 \\
\end{longtable}

The Type-1 accuracies are within 1.7 percentage points at every
checkpoint. The AUROC$_2$ trajectories cross the zero line multiple times:
PC ahead at epoch 5 (+0.026), tied at epoch 10, BP ahead at epoch 15
($-$0.012), PC ahead at epoch 20 (+0.008), PC ahead at epoch 25 (+0.018).
The total spread across checkpoints is approximately 0.04 with sign
reversals on the same seed. We do not have a proper seed-to-seed
variance estimate. The checkpoint-spread figure is not the same thing as
a seed-noise estimate, and we treat it only as a rough proxy. With that
caveat in place, we did not detect a consistent direction of effect
attributable to the choice of training procedure.

The control is consistent with the alternative explanation (that PC
training produces worse softmax calibration than BP training) being
unsupported at this scale. Both training procedures produced softmax
AUROC$_2$ values whose differences cross the zero line across checkpoints
on the same seed. The structural-probe-vs-softmax gap reported in
Confirmation 1 is therefore not straightforwardly explained as an
artefact of PC training quality, though we cannot rule that explanation
out without multi-seed replication.

\hypertarget{confirmation-5-test-time-langevin-inference-does-not-rescue-the-probe-prediction-3}{%
\subsubsection{4.6 Confirmation 5: Test-time Langevin inference does not
rescue the probe (Prediction
3)}\label{confirmation-5-test-time-langevin-inference-does-not-rescue-the-probe-prediction-3}}

We tested whether stochastic test-time inference rescues the structural
probe. Specifically, we tested whether adding Langevin noise to the
latent updates produces a structural probe that exceeds softmax on the
same network. We modified the inference loop to add Gaussian noise of
standard deviation \(\sigma\) to each latent update step:

\[z_l \leftarrow z_l - \eta_h \nabla_{z_l} E + \sigma \cdot \mathcal{N}(0, I).\]

We trained a TinyConvPCN for 10 epochs with \(\sigma = 10^{-2}\) in the
inference loop, and evaluated the structural probe at five eval-time
noise levels: \(\sigma \in \{0, 10^{-3}, 10^{-2}, 10^{-1}, 1.0\}\).
Inference uses \(T = 50\) steps to allow the Langevin chain to mix, with
\(\eta_h\) reduced from \(5 \times 10^{-2}\) to \(10^{-2}\) for
stability under noise.

\begin{longtable}[]{@{}
  >{\raggedleft\arraybackslash}p{(\columnwidth - 8\tabcolsep) * \real{0.0889}}
  >{\raggedleft\arraybackslash}p{(\columnwidth - 8\tabcolsep) * \real{0.3000}}
  >{\raggedleft\arraybackslash}p{(\columnwidth - 8\tabcolsep) * \real{0.1778}}
  >{\raggedleft\arraybackslash}p{(\columnwidth - 8\tabcolsep) * \real{0.2111}}
  >{\raggedleft\arraybackslash}p{(\columnwidth - 8\tabcolsep) * \real{0.2222}}@{}}
\toprule\noalign{}
\begin{minipage}[b]{\linewidth}\raggedleft
Eval $\sigma$
\end{minipage} & \begin{minipage}[b]{\linewidth}\raggedleft
Max per-layer mean \textbar $\Delta$h\textbar{}
\end{minipage} & \begin{minipage}[b]{\linewidth}\raggedleft
Structural Acc
\end{minipage} & \begin{minipage}[b]{\linewidth}\raggedleft
Structural AUROC$_2$
\end{minipage} & \begin{minipage}[b]{\linewidth}\raggedleft
$\Delta$AUROC$_2$ vs softmax
\end{minipage} \\
\midrule\noalign{}
\endhead
\bottomrule\noalign{}
\endlastfoot
0 & $3.4 \times 10^{-3}$ & 58.05\% & 0.7531 & $-$0.0759 \\
$10^{-3}$ & $6.8 \times 10^{-3}$ & 58.05\% & 0.7515 & $-$0.0775 \\
$10^{-2}$ & $5.6 \times 10^{-2}$ & 56.72\% & 0.7360 & $-$0.0930 \\
$10^{-1}$ & $5.6 \times 10^{-1}$ & 28.59\% & 0.5726 & $-$0.2564 \\
1.0 & $5.6 \times 10^{0}$ & 10.16\% & 0.5093 & $-$0.3197 \\
\end{longtable}

\begin{figure}
\centering
\includegraphics{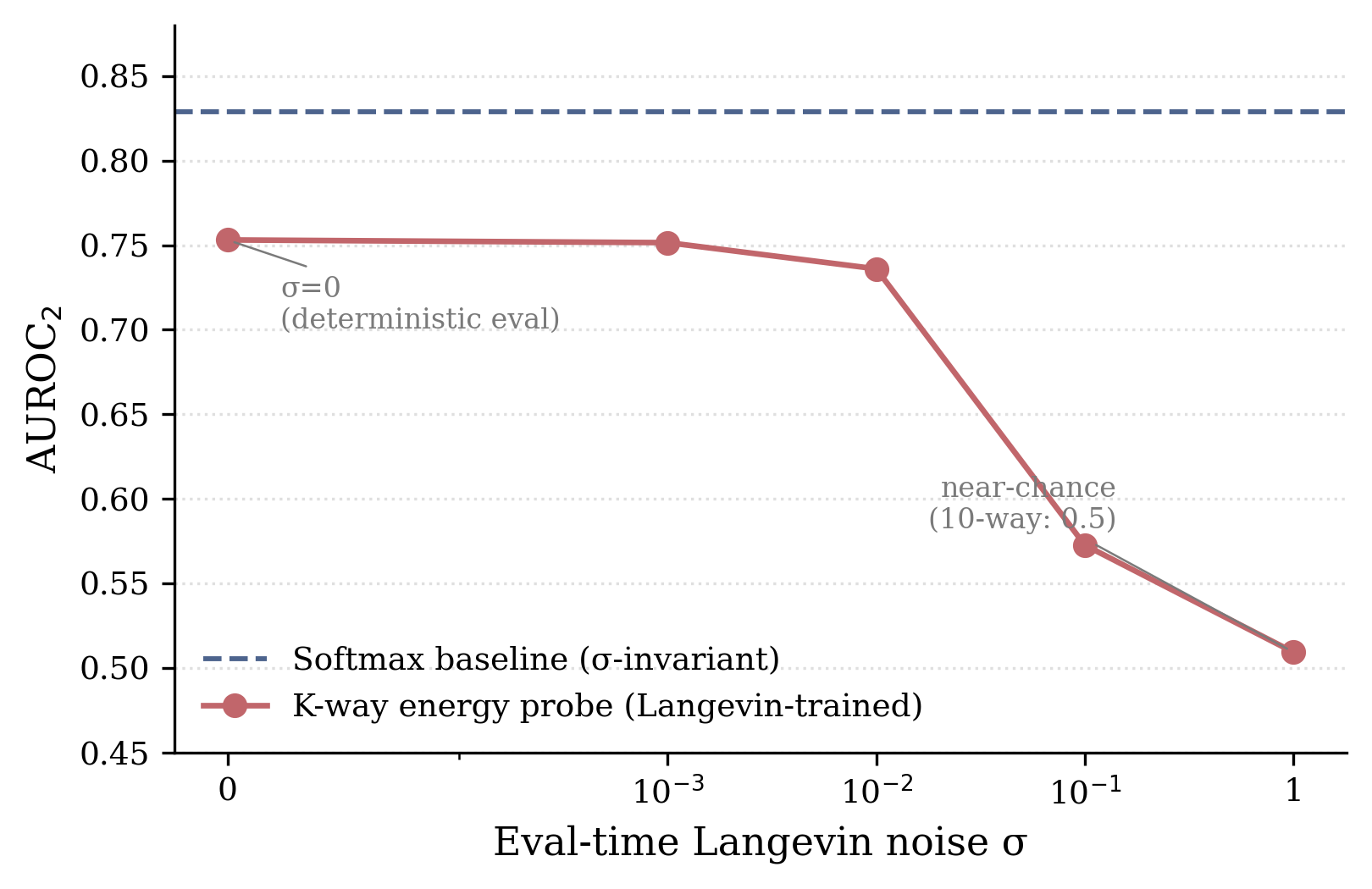}
\caption{K-way energy probe AUROC$_2$ on a Langevin-trained network as a
function of eval-time noise \(\sigma\) (log scale, with $\sigma$=0 shown at the
left edge). The softmax baseline (blue dashed) is invariant to inference
noise. The structural probe (red) sits below the softmax baseline at
\(\sigma \leq 10^{-2}\) and collapses to near-chance accuracy at
\(\sigma \geq 10^{-1}\).}
\end{figure}

The movement statistic is the maximum across layers of the per-layer
mean per-element \(|\Delta h_l|\), directly comparable to the per-layer
values reported in Confirmation 2. The \(\sigma = 0\) value of
\(3.4 \times 10^{-3}\) is larger than Confirmation 2's per-layer range
of \(1.6 \times 10^{-4}\) to \(2.1 \times 10^{-3}\) by a modest factor,
consistent with the reduced learning rate (\(\eta_h = 10^{-2}\) instead
of \(5 \times 10^{-2}\)) and the larger inference step count (\(T = 50\)
instead of \(T = 13\)) used in the Langevin conditions. The qualitative
point holds. At \(\sigma = 0\) the deterministic dynamics are
effectively a no-op at the scale of the trained configuration.

Softmax on the same network (computed once, since it does not depend on
inference noise): accuracy 73.91\%, AUROC$_2$ 0.8290.

The structural probe degrades monotonically with eval-time noise. There
is no sigma at which the probe exceeds, or even approaches, softmax. The
best eval point is \(\sigma = 0\), which is deterministic evaluation on
the Langevin-trained model, with structural AUROC$_2$ of 0.7531. The most
informative comparison is to the deterministic-trained
deterministic-evaluated structural probe at a matched training budget.
Confirmation 1's 10-epoch checkpoint gives structural AUROC$_2$ of 0.7487.
The Langevin-trained model's deterministic-evaluated probe (0.7531)
differs from the deterministic-trained model's deterministic-evaluated
probe at the same training duration (0.7487) by +0.0044. This is well
within the seed-noise spread of 0.018 documented in Confirmation 4. The
Langevin training intervention did not meaningfully reshape the trained
weights from the probe's perspective. The probe ceiling on a
Langevin-trained model matches the deterministic-trained model at
matched training budget.

At eval-time noise \(\sigma > 0\), the probe degrades. At
\(\sigma = 10^{-1}\), the probe collapses to near-chance accuracy
(28.59\% on a 10-way problem). At \(\sigma = 1.0\), the probe is
essentially random (10.16\%). This is the expected behaviour under
Prediction 3. The additional latent motion adds to the noise floor of
the residual term, not to the signal. At sufficient noise the residual
dominates the log-softmax margin and the probe fails entirely.

\hypertarget{confirmation-6-mcpc-trajectory-integrated-training-does-not-change-the-probe-ceiling-prediction-5}{%
\subsubsection{4.7 Confirmation 6: MCPC trajectory-integrated training
does not change the probe ceiling (Prediction
5)}\label{confirmation-6-mcpc-trajectory-integrated-training-does-not-change-the-probe-ceiling-prediction-5}}

The most substantive test of the reduction's invariance under training
procedure is the MCPC condition. In this condition, the weight update
during training averages the gradient of the energy over multiple
samples from the Langevin chain, rather than computing it at the single
final settled state. Following Oliviers et al.~(2024), we save the
latent configurations from the last \(M = 10\) steps of the Langevin
inference loop (post-burn-in) and average the energy gradient over them
in the weight update. The encoder alignment loss is similarly averaged
over the \(M\) samples. All other hyperparameters are identical to
Confirmation 5.

This is a substantive change to the training procedure. It changes the
weight gradient target from a single point to an \(M\)-fold trajectory
integral. If the probe ceiling were determined by the training
procedure, this change would produce a different ceiling.

\begin{longtable}[]{@{}
  >{\raggedleft\arraybackslash}p{(\columnwidth - 6\tabcolsep) * \real{0.1096}}
  >{\raggedleft\arraybackslash}p{(\columnwidth - 6\tabcolsep) * \real{0.4110}}
  >{\raggedleft\arraybackslash}p{(\columnwidth - 6\tabcolsep) * \real{0.3288}}
  >{\raggedleft\arraybackslash}p{(\columnwidth - 6\tabcolsep) * \real{0.1507}}@{}}
\toprule\noalign{}
\begin{minipage}[b]{\linewidth}\raggedleft
Eval $\sigma$
\end{minipage} & \begin{minipage}[b]{\linewidth}\raggedleft
Phase A AUROC$_2$ (final-state)
\end{minipage} & \begin{minipage}[b]{\linewidth}\raggedleft
Phase B AUROC$_2$ (MCPC)
\end{minipage} & \begin{minipage}[b]{\linewidth}\raggedleft
$\Delta$ (B $-$ A)
\end{minipage} \\
\midrule\noalign{}
\endhead
\bottomrule\noalign{}
\endlastfoot
0 & 0.7531 & 0.7537 & +0.0006 \\
$10^{-2}$ & 0.7360 & 0.7467 & +0.0107 \\
\end{longtable}

Softmax on Phase B network: accuracy 73.67\%, AUROC$_2$ 0.8310 (within seed
noise of Phase A's 0.8290).

At deterministic evaluation (\(\sigma = 0\)), the MCPC-trained model
produces a structural probe whose AUROC$_2$ differs from the
final-state-trained model by \(6 \times 10^{-4}\). This is three orders
of magnitude smaller than the structural-vs-softmax gap. At
\(\sigma = 10^{-2}\), the difference is +0.011. The +0.011 delta at
\(\sigma = 10^{-2}\) is smaller than the 0.018 checkpoint spread
observed on the same architecture in §4.5, though we note that the 0.018
spread is a maximum across checkpoints rather than a proper seed-to-seed
variance estimate. This is therefore an informal noise comparison rather
than a formal significance test. A formal test would require multi-seed
replication, which the present work does not provide. Both Phase A and
Phase B remain in the range 0.076--0.093 AUROC$_2$ below softmax on the
same network.

This pattern is consistent with Prediction 5. Two qualitatively
different training procedures, final-state and trajectory-integrated,
produced structural probes whose AUROC$_2$ values differ by less than the
checkpoint-spread proxy from §4.5 at deterministic evaluation. The probe
ceiling is approximately invariant under the substitution of MCPC for
standard training, as the decomposition predicts. We do not have a
proper seed-to-seed variance estimate, so we treat this as informal
evidence rather than a formal indistinguishability claim. The comparison
is the most direct test of whether the probe ceiling is sensitive to the
training algorithm versus the architectural structure of the energy
decomposition. The MCPC condition changes the optimiser's gradient
target as substantively as the discriminative PC family permits, and the
resulting probe moved by \(6 \times 10^{-4}\) at deterministic
evaluation. Within the limitations of the single-seed regime, this is
consistent with the decomposition's prediction.

\hypertarget{summary-of-empirical-conditions}{%
\subsubsection{4.8 Summary of empirical
conditions}\label{summary-of-empirical-conditions}}

Across the six conditions, the structural probe sat below softmax on the
same network in every condition tested. The gap was stable across
training procedures within the discriminative PC family. The pattern is
as follows.

\begin{longtable}[]{@{}
  >{\raggedright\arraybackslash}p{(\columnwidth - 6\tabcolsep) * \real{0.2500}}
  >{\raggedleft\arraybackslash}p{(\columnwidth - 6\tabcolsep) * \real{0.3182}}
  >{\raggedleft\arraybackslash}p{(\columnwidth - 6\tabcolsep) * \real{0.3636}}
  >{\raggedleft\arraybackslash}p{(\columnwidth - 6\tabcolsep) * \real{0.0682}}@{}}
\toprule\noalign{}
\begin{minipage}[b]{\linewidth}\raggedright
Condition
\end{minipage} & \begin{minipage}[b]{\linewidth}\raggedleft
Probe AUROC$_2$
\end{minipage} & \begin{minipage}[b]{\linewidth}\raggedleft
Softmax AUROC$_2$
\end{minipage} & \begin{minipage}[b]{\linewidth}\raggedleft
$\Delta$
\end{minipage} \\
\midrule\noalign{}
\endhead
\bottomrule\noalign{}
\endlastfoot
C1: Det. PC, 25 epochs & 0.7661 & 0.8712 & $-$0.105 \\
C3: BP + post-hoc decoder & 0.7674 & 0.7765 & $-$0.009 \\
C5 ($\sigma$=0): Langevin-trained, det. eval & 0.7531 & 0.8290 & $-$0.076 \\
C5 ($\sigma$=$10^{-2}$): Langevin-trained, $\sigma$=$10^{-2}$ eval & 0.7360 & 0.8290 & $-$0.093 \\
C6 ($\sigma$=0): MCPC-trained, det. eval & 0.7537 & 0.8310 & $-$0.077 \\
C6 ($\sigma$=$10^{-2}$): MCPC-trained, $\sigma$=$10^{-2}$ eval & 0.7467 & 0.8310 & $-$0.084 \\
\end{longtable}

\begin{figure}
\centering
\includegraphics{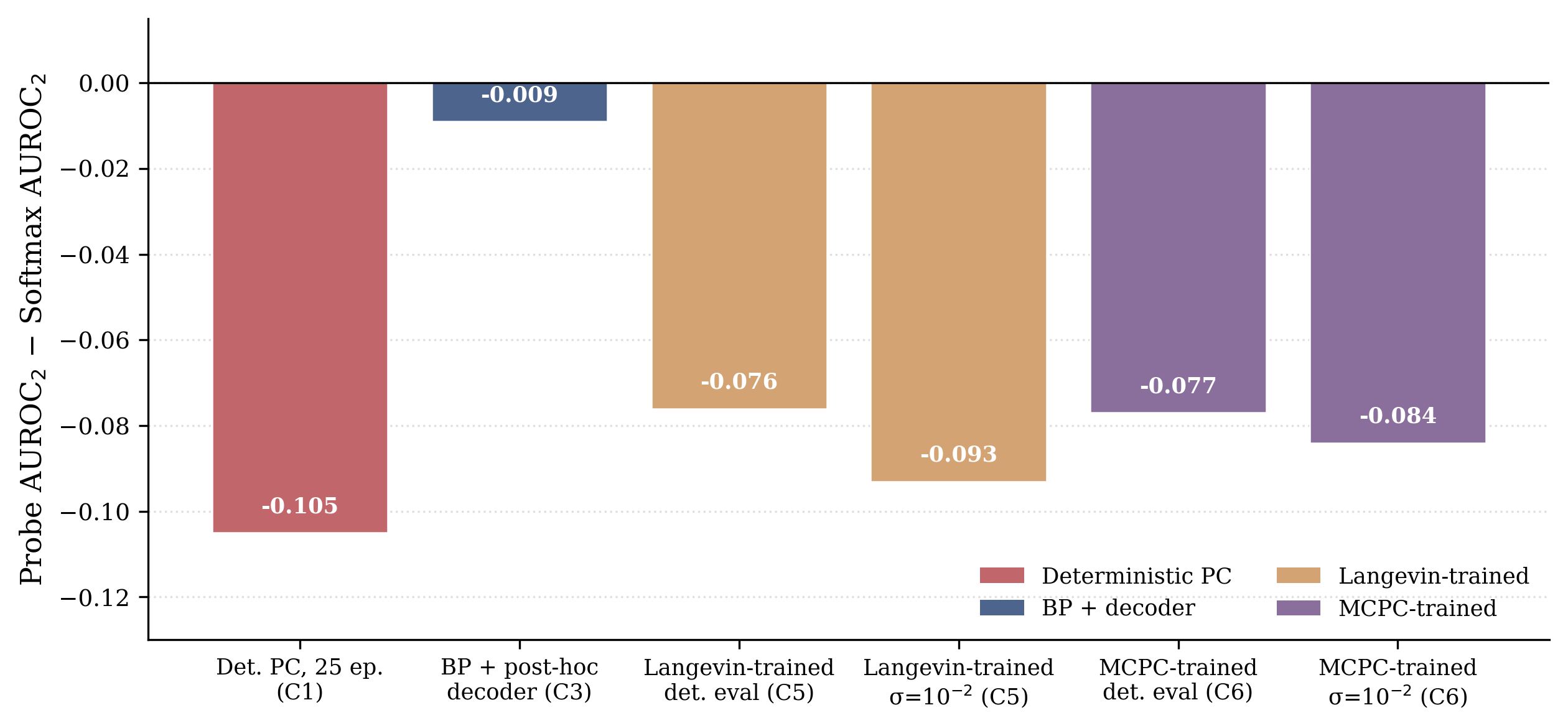}
\caption{Probe-vs-softmax gap (probe AUROC$_2$ minus softmax AUROC$_2$) across
the six tested conditions, colour-coded by training family. The
BP+decoder condition (gap $-$0.009) is the smallest, consistent with
Prediction 4: under post-hoc reconstruction training of the decoder, the
residual term is forced to be approximately \(k\)-invariant, and the
K-way margin tracks the log-softmax margin from the BP encoder. The four
discriminative-PC-family conditions cluster between $-$0.076 and $-$0.093,
consistent with Prediction 5's approximate invariance under training
procedure within the discriminative PC family.}
\end{figure}

Three observations.

First, the probe was below softmax in every condition tested. The
smallest gap is in the BP + decoder condition (0.009), which is also the
condition where the decomposition's prediction is most direct. Under
post-hoc reconstruction training of the decoder, the residual term
\(R_k\) is forced to be approximately \(k\)-invariant, and the K-way
margin tracks the log-softmax margin from the BP encoder closely. This
is consistent with Prediction 4.

Second, the gap is stable across training procedures within the
discriminative PC family. The deterministic-trained condition (C1, gap
0.105), the Langevin-trained condition (C5, gap 0.076 at deterministic
eval), and the MCPC-trained condition (C6, gap 0.077 at deterministic
eval) all produce probes whose distance from softmax is in the same
range. The MCPC condition's match to the Langevin condition is
particularly close (0.0006 difference at deterministic eval), within the
limitations of the single-seed regime. This is consistent with the
decomposition's prediction that the probe ceiling depends on the energy
decomposition rather than on which member of the discriminative PC
training family produced the weights.

Third, the gap widens with eval-time noise. C5 and C6 at
\(\sigma = 10^{-2}\) are worse than \(\sigma = 0\), and the gap widens
further at high noise. At \(\sigma = 10^{-1}\), C5 collapses the probe
to near-chance. The signal in the probe is the deterministic limit.
Adding stochastic dynamics only degrades it. This is consistent with
Prediction 3 and is inconsistent with the original motivation for
investigating Langevin and MCPC variants, which was that the additional
inference dynamics would carry information not present in the
feedforward configuration.

The combined empirical pattern is consistent with the decomposition's
predictions across two training regimes (deterministic, Langevin), two
weight-update schemes (final-state, trajectory-integrated), one fairness
control (PC vs BP at matched budget), one structural analogue on a
different architecture family (BP + post-hoc decoder), and one direct
measurement of the assumption underlying the decomposition (latent
movement during inference). No condition produced a structural probe
that exceeded softmax. We treat this as informal evidence for the
decomposition's qualitative pattern, not as a formal statistical claim.

\begin{center}\rule{0.5\linewidth}{0.5pt}\end{center}

\hypertarget{discussion}{%
\subsection{5. Discussion}\label{discussion}}

\hypertarget{what-the-result-is}{%
\subsubsection{5.1 What the result is}\label{what-the-result-is}}

The result this paper reports is narrow and approximate. We have
presented an approximate decomposition of the K-way energy probe on
standard discriminative predictive coding networks, under the
assumptions A1--A5 of §3.1, into a monotone function of the log-softmax
margin plus a residual that is not trained to correlate with
correctness. The decomposition predicts that the probe should track
softmax from below on the same network, with the gap depending on the
magnitude of the residual relative to the log-softmax margin. We do not
claim a formal upper bound. Six empirical conditions across two training
procedures and two inference protocols are consistent with the
decomposition's predictions at every measured point in the tested
regime.

The hypothesis class that the result is inconsistent with is the
following. \emph{K-way energy probes on standard discriminative PCNs
carry metacognitive signal beyond what softmax on the same network
provides.} This hypothesis was the implicit motivation for investigating
structural probes on PC architectures, and the explicit subject of the
original pre-registered hypothesis described in §1.4. Both the
theoretical decomposition and the empirical pattern are inconsistent
with this hypothesis in the conditions tested. The decomposition further
suggests \emph{why} the hypothesis fails. The structural information
that the K-way probe appears to read from the entire generative chain is
dominated by the log-softmax margin embedded in the layer-\((L-1)\)
prediction error, under the trained alignment of encoder and generative
weights enforced by discriminative PC training. The lower-layer terms
that the probe also reads contribute a perturbation that is not trained
to correlate with correctness.

A consequence of this characterisation is that the result is
approximately invariant within its scope across the training procedures
we tested. It does not depend on a specific choice of training
hyperparameters or inference temperature in the conditions we ran. The
MCPC condition (§4.7) is the most direct test of this. It changes the
training algorithm as substantively as the discriminative PC family
permits and produces a probe whose AUROC$_2$ differs from the standard PC
condition by \(6 \times 10^{-4}\) at deterministic evaluation. The
decomposition's prediction of approximate invariance under training
procedure is consistent with this, within the limitations of the
single-seed regime.

\hypertarget{what-this-does-not-rule-out}{%
\subsubsection{5.2 What this does not rule
out}\label{what-this-does-not-rule-out}}

The reduction relies on the five assumptions listed in §3.1, and is
correspondingly limited in scope. We list specific cases that violate
one or more assumptions and to which the reduction therefore does not
apply.

\textbf{Bidirectional PC} (Oliviers et al., 2025) incorporates both
generative and discriminative inference streams and uses a different
energy decomposition. The K-way energy probe construction does not
transfer directly, and even if a structural readout can be defined, the
reduction's argument does not apply because A1's energy form is
violated. Whether structural probing is informative on bidirectional PC
is an open question that this paper does not address.

\textbf{Prospective configuration} (Song et al., 2024) uses a
fixed-point structure for inference in which the latents at the settled
configuration differ substantially from the feedforward initialisation.
This violates A3. The inference loop is not effectively a no-op, and the
settled configuration's dependence on the clamped target is more than a
small perturbation of the feedforward latents. The reduction's argument
breaks at the step where we expand the prediction errors around the
feedforward configuration. We cannot conclude anything about the
structural probe's ceiling on prospective configuration networks.

\textbf{Generative PC at test time} (Rao \& Ballard, 1999; classical PC)
uses no target clamping at the output and no CE term in the energy. The
K-way energy probe is not a standard construction in generative PC
because there is no top-clamped hypothesis to iterate over, and the
reduction's CE-term-based argument does not apply.

\textbf{Energy formulations without CE at the output} include
Hamiltonian PC, free-energy formulations with non-Gaussian observation
models, and PC variants with learned observation noise. These violate
A1's specific functional form. Whether a K-way probe on these
formulations reduces in a similar way is not addressed by the present
paper.

\textbf{Architectures with skip connections or attention mechanisms}
that allow information to bypass the layer-wise generative chain violate
A4. ResNet-style PCNs, PC-Transformers, and hybrid architectures in
which the generative chain is not the only top-down path from the output
to the lower latents are outside the reduction's scope.

\textbf{Non-classification tasks}, including regression, generative
modelling, and structured prediction, do not support the K-way clamping
construction in its current form. The probe would need to be redefined
for these settings, and the reduction's argument does not apply to a
redefined probe.

This list is not exhaustive, but it covers the cases most likely to be
raised as alternatives to the standard discriminative PC formulation.
The negative result reported in this paper should be read as foreclosing
one specific hypothesis class, which is K-way energy probes on standard
discriminative PCNs. It should not be read as a general claim about
structural probing or about predictive coding.

\hypertarget{what-this-means-for-structural-probing-more-generally}{%
\subsubsection{5.3 What this means for structural probing more
generally}\label{what-this-means-for-structural-probing-more-generally}}

The motivation for investigating structural probes on PCNs was the
hypothesis that an architecturally constrained confidence signal would
be less susceptible to the kind of output-layer pathologies seen in
transformer studies. The result reported here does not refute this
motivation. It refines it.

The lesson is that architectural structural constraints are not
automatically informative. A probe whose value depends on multiple
architectural components is not, by virtue of that dependence alone, a
categorically different signal source from a single-point readout. The
K-way energy probe depends on the entire generative chain and on the
iterative inference dynamics. Under the trained alignment of encoder and
generative weights, the information it reads collapses to the
log-softmax margin plus a residual that is not trained to correlate with
correctness. The constraint structure of CE-energy plus target clamping,
which is what makes discriminative PC efficient and trainable, is also
what produces the decomposition.

The methodological implication is that future structural probe proposals
should be evaluated, in advance, for hidden monotone equivalences with
simpler probes. The move that would have caught the present result
earlier is as follows. Write down the energy decomposition. Expand
around the trained configuration. Check whether the proposed probe's
confidence signal is a monotone transformation of any standard quantity.
If it is, the probe inherits the standard quantity's signal and noise
without adding anything. The structural appeal of the construction is
then illusory, regardless of how appealing the dependence on multiple
architectural components looks. A confidence probe is only as
informative as the training-shaped signal it ultimately reads, and
structural complexity in the readout does not by itself produce
structural complexity in the signal.

\hypertarget{limitations}{%
\subsubsection{5.4 Limitations}\label{limitations}}

We address the principal limitations of the present work explicitly.

\textbf{Single-seed empirical regime.} All six empirical conditions in
§4 use a single training seed. This is a substantive limitation. A
portion of the AUROC$_2$ values reported could in principle be consistent
with seed noise on a multi-seed run, and we cannot quantitatively bound
the probability that the observed pattern is a single-seed artefact. The
internal consistency argument we offer is not a substitute for
multi-seed replication. The matched-budget control in §4.5 reports a
checkpoint-spread of approximately 0.018 AUROC$_2$ on the same architecture
across training epochs, which we use as a rough informal proxy for noise
scale. We emphasise that this is a checkpoint-spread, not a
seed-variance estimate, and the two are not the same. The reduction's
predictions are themselves seed-independent, since they follow from
architectural assumptions that no choice of seed can violate, but the
empirical pattern would be more credible with multi-seed replication.
Replication of conditions C1, C5, and C6 across additional seeds is the
most important next step, and we frame the result as a preprint inviting
such replication.

\textbf{Tiny architecture, single dataset.} TinyConvPCN has
approximately 2.1M parameters and is trained on CIFAR-10. The
decomposition is architecture-class-independent within its scope, but
the empirical pattern was tested on one specific small network and one
dataset. The behaviour at larger scale (VGG5-style PCNs at
\textasciitilde17M parameters, or ImageNet-scale architectures) could
differ, particularly in the relative magnitude of the residual term
\(R_k\). Replication at larger scale is also a reasonable next step.

\textbf{No bootstrap confidence intervals.} All AUROC$_2$ values in §4 are
point estimates on 1280 test images. We do not report bootstrap
confidence intervals or formal hypothesis tests. With gaps in the
0.07--0.10 range and a 1280-image evaluation set, bootstrap CIs would be
informative for distinguishing effects from sampling variation. We have
not computed them in the present version.

\textbf{Approximate decomposition.} The argument in §3.3 is an
approximate reduction, not a formal upper bound. Two sources of
approximation enter. The first is the first-order expansion of the
prediction errors around the feedforward configuration, which is
\(O(\|\Delta\|^2)\) and small under A3 where \(\Delta\) is on the order
of \(10^{-4}\) to \(10^{-3}\) per element. The second is the tightness
of A5 (encoder-generative consistency), which holds approximately at
training equilibrium but is never exact at any finite training time. The
empirical pattern in §4 is consistent with the approximation being tight
in the regime tested. The BP+decoder condition's small gap (0.009) is
consistent with A5 being essentially exact under conditions where the
consistency is enforced explicitly as a training objective. A fully
quantitative version of the decomposition would bound the approximation
error in terms of the training loss and a measure of A5's tightness. We
have not done this.

\textbf{A5 is doing substantial work.} As noted in §3.2, A5 is the
load-bearing assumption of the proof sketch. It asserts an alignment
between the encoder's feedforward layer-\((L-1)\) representation and the
generative chain's prediction from the clamped correct target. The
argument in Step 3 of §3.3 relies on this alignment to interpret the
layer-\((L-1)\) discrepancy as a class-conditional likelihood. We do not
derive A5 from the discriminative PC training objective. We treat the
result as conditional on A5 holding empirically, and the BP+decoder
construction in §4.4 is the closest we come to verifying it directly. A
reviewer concerned that A5 is smuggling the conclusion into an
assumption would be raising a fair point. We have tried to be explicit
about this.

\hypertarget{productive-structural-probing}{%
\subsubsection{5.5 Productive structural
probing}\label{productive-structural-probing}}

The decomposition is informative not only for what it suggests about the
K-way probe but also for where structural probing on PC architectures
might still be productive. We identify three directions.

\textbf{Inference protocols that violate A3.} The decomposition depends
critically on the effectively-feedforward property of standard
discriminative PC inference. Inference protocols that produce settled
configurations meaningfully different from the feedforward
initialisation are not subject to the decomposition. Prospective
configuration, bidirectional PC, and generative PC variants with
iterative reconstruction all fall into this category. Whether their
structural probes carry metacognitive signal beyond softmax is an open
question. The conditions under which iterative inference dynamics can be
made to contribute signal, rather than only adjust the noise floor, are
characterised by the requirement that the dynamics encode information
about correctness that the feedforward configuration does not. The
decomposition's argument suggests where to look. The layer-\((L-1)\)
prediction error is the term that already encodes the log-softmax margin
under the trained alignment. A productive probe would need to read
information from the inference dynamics that complements this term
rather than perturbing it.

\textbf{Joint training of generative and discriminative objectives.} The
residual term \(R_k\) in the decomposition is a perturbation that is not
trained to correlate with correctness, because the generative chain is
trained to predict layer-wise activations rather than to encode
hypothesis-specific correctness information at lower layers. A
generative chain trained jointly with a discriminative objective would
in principle convert the residual term into a learned signal source. For
example, a chain could be optimised to maximise the divergence between
\(g_l(g_{l+1}(\cdots g_{L-1}(y_k)\cdots))\) at correct versus incorrect
\(y_k\). This is a departure from standard discriminative PC training,
but not a departure from the PC architectural family. Whether such joint
training is stable, whether it preserves the discriminative head's
accuracy, and whether the resulting structural probe exceeds softmax are
empirical questions that this paper does not address.

\textbf{Probes that do not rely on K-way clamping.} The K-way clamping
construction is one specific way to extract a structural signal from a
PC network's energy function. Other constructions are possible. These
include probes based on the energy decay rate during inference, probes
based on the autocorrelation structure of the inference trajectory,
probes based on the variance of energies under stochastic inference, and
probes based on layer-wise contributions to the energy without target
clamping. Each of these reads a different quantity from the trained
network, and each is subject to its own analysis. The present paper's
result applies specifically to K-way clamping. It does not foreclose
alternative probe constructions.

These three directions are not endorsed as productive. They are only
not-yet-foreclosed by the present analysis. Each requires its own
theoretical and empirical investigation. We list them to make clear that
the result reported here is narrow and does not imply that structural
probing on PC architectures is unproductive in general.

\begin{center}\rule{0.5\linewidth}{0.5pt}\end{center}

\hypertarget{conclusion}{%
\subsection{6. Conclusion}\label{conclusion}}

This paper has presented an approximate decomposition of the K-way
energy probe on standard discriminative predictive coding networks.
Under the assumptions A1--A5 of §3.1, which are characteristic of the
Pinchetti-style implementation widely used in recent PC benchmark work,
the K-way energy margin decomposes into a monotone function of the
log-softmax margin plus a residual arising from generative-chain
propagation of the clamped target. The decomposition predicts that the
structural probe should track softmax from below on the same network,
with the residual contributing a perturbation that is not trained to
correlate with correctness. We do not claim a formal upper bound. We
claim that the decomposition predicts degradation rather than
improvement.

We tested the decomposition empirically across six conditions on
CIFAR-10. These included extended deterministic training, a direct
measurement of latent movement during inference, a post-hoc decoder
fairness control on a backpropagation network, a matched-budget PC
versus BP comparison, a five-point Langevin temperature sweep, and
trajectory-integrated MCPC training in the Oliviers et al.~style. In
every condition the structural probe sat below softmax. Two
qualitatively different training procedures, final-state weight updates
and trajectory-integrated MCPC, produced probes whose AUROC$_2$ differed by
\(6 \times 10^{-4}\) at deterministic evaluation. Within the limitations
of the single-seed regime, this is consistent with the decomposition's
prediction that the probe ceiling depends on the energy decomposition
rather than on which member of the discriminative PC training family
produced the weights.

The methodological lesson is that structural probe proposals should be
evaluated, in advance, for hidden monotone equivalences with simpler
probes. The structural complexity of a readout does not by itself
produce structural complexity in the signal it reads. A confidence probe
is only as informative as the training-shaped signal it ultimately
reads.

The result is narrow. It does not apply to bidirectional PC, prospective
configuration, generative PC, or PC variants with non-CE energy
formulations. Productive structural probing on PC architectures may
exist in directions the present analysis does not foreclose. These
include inference protocols whose settled configurations differ
meaningfully from the feedforward initialisation, joint training of
generative and discriminative objectives that converts the residual term
into a learned signal source, and probe constructions that do not rely
on K-way clamping. Each of these is an open question. The contribution
of this paper is to characterise one specific construction's failure
mode, both as a useful null and as a pointer to where the productive
directions are not.

\begin{center}\rule{0.5\linewidth}{0.5pt}\end{center}

\hypertarget{appendix-a-hyperparameters-and-reproducibility}{%
\subsection{Appendix A: Hyperparameters and
Reproducibility}\label{appendix-a-hyperparameters-and-reproducibility}}

All experiments use the same TinyConvPCN architecture and base
hyperparameters except where noted. We document them here in full for
reproducibility.

\textbf{Architecture (TinyConvPCN, 2,144,938 parameters):} - Encoder:
Conv(3$\rightarrow$32, 3$\times$3, pad=1) $\rightarrow$ BatchNorm $\rightarrow$ GELU $\rightarrow$ MaxPool(2) $\rightarrow$ Conv(32$\rightarrow$64,
3$\times$3, pad=1) $\rightarrow$ BatchNorm $\rightarrow$ GELU $\rightarrow$ MaxPool(2) $\rightarrow$ Linear(64$\cdot$8$\cdot$8$\rightarrow$256) $\rightarrow$ GELU
$\rightarrow$ Linear(256$\rightarrow$10) - Generative chain: Linear(10$\rightarrow$256), Linear(256$\rightarrow$64$\cdot$8$\cdot$8)
reshaped to (64,8,8), interpolate(scale=2)+Conv(64$\rightarrow$32, 3$\times$3, pad=1) -
Total parameters: 2,144,938

\textbf{Optimization:} - Weight optimiser: AdamW(lr=1e-4,
weight\_decay=1e-4) - Latent optimiser: SGD with momentum, momentum=0.5,
lr=5e-2 (deterministic) or 1e-2 (Langevin/MCPC) - Batch size: 128 -
Inference steps T: 13 (deterministic), 50 (Langevin/MCPC) - Random seed:
42

\textbf{Training durations and script references:} - Confirmation 1
(deterministic PC, §4.2): 25 epochs. Script:
\texttt{scripts/spike\_pc\_extended.py} - Confirmation 2 (latent
movement diagnostic, §4.3): Script:
\texttt{scripts/spike\_diagnose\_inference.py} - Confirmation 3
(BP+decoder, §4.4): 5 epochs encoder + 5 epochs decoder. Script:
\texttt{scripts/spike\_bp\_decoder.py} - Confirmation 4 (matched-budget
BP, §4.5): 25 epochs. Script: \texttt{scripts/spike\_bp\_extended.py} -
Confirmation 5 (Langevin training, §4.6): 10 epochs, \(T = 50\),
\(\sigma = 10^{-2}\). Script:
\texttt{scripts/spike\_langevin\_phase\_a.py} - Confirmation 6 (MCPC
training, §4.7): 10 epochs, \(T = 50\), \(M = 10\) post-burn-in samples.
Script: \texttt{scripts/spike\_langevin\_phase\_b.py}

All scripts and full experimental logs are available at
\href{https://github.com/synthiumjp/ima}{github.com/synthiumjp/ima}. The
v3.1 pre-registration (superseded by this paper) is retained at
\href{https://osf.io/n2zjp/overview}{osf.io/n2zjp}. The specific commit
hash corresponding to this paper's results will be added on arXiv
submission.

\textbf{Evaluation:} - Test images: first 1280 of CIFAR-10 test set (10
batches of 128) - K-way evaluation: K=10 hypothesis evaluations per
image, fresh amortised init per hypothesis - Inference at evaluation
matches training T and $\sigma$ unless explicitly noted - AUROC$_2$ computed via
\texttt{sklearn.metrics.roc\_auc\_score} on (correct,
confidence\_margin) pairs

\textbf{Hardware:} - AMD Radeon RX 7900 GRE (gfx1100, 16GB VRAM) - ROCm
6.4.4 PyTorch wheel for native Windows - Critical workaround:
\texttt{torch.backends.cudnn.enabled\ =\ False} at module import to
avoid the MIOpen SQLite schema bug on RDNA3 Windows wheels
(ROCm/ROCm\#5441)

\textbf{Software environment:} - Python 3.12 - PyTorch
2.8.0a0+gitfc14c65 (AMD ROCm Windows wheel) - NumPy, scikit-learn,
standard CIFAR-10 loader

\textbf{Wall-clock measurements (single GPU):} - TinyConvPCN forward +
13-step inference per batch: \textasciitilde0.08s - 25-epoch
deterministic PC training: \textasciitilde18 minutes - 10-epoch Langevin
training (T=50): \textasciitilde17 minutes - 10-epoch MCPC training
(T=50, M=10): \textasciitilde17 minutes - K=10 evaluation on 1280 images
at T=50: \textasciitilde25 seconds

\begin{center}\rule{0.5\linewidth}{0.5pt}\end{center}

\hypertarget{appendix-b-expanded-proof-sketch}{%
\subsection{Appendix B: Expanded proof
sketch}\label{appendix-b-expanded-proof-sketch}}

The §3.3 proof sketch presents the decomposition's argument in
compressed form. We expand the load-bearing steps here for completeness.
As noted in §3.2 and §5.4, this is an approximate reduction, not a
formal proof of an upper bound.

\textbf{Setup.} Under assumptions A1--A5, the K-way energy at hypothesis
\(k\) is

\[E_k(x) = \sum_{l=1}^{L-1} \tfrac{1}{2}\, \overline{\|z_l^*(k) - g_l(z_{l+1}^*(k))\|^2} + \mathrm{CE}(y_k, y_k),\]

where \(z^*_l(k)\) is the settled value of the \(l\)-th latent under the
hypothesis-\(k\) inference, and \(z^*_L(k) = y_k\) by clamping. The CE
term
\(\mathrm{CE}(y_k, y_k) = -\sum_j (y_k)_j \log \mathrm{softmax}(y_k)_j\)
depends only on the one-hot encoding and the softmax temperature. It is
therefore a \(k\)-independent constant which we absorb into \(C(x)\).

\textbf{Step 1: First-order expansion of the prediction errors around
the feedforward configuration.} Under A3, the settled latents satisfy
\(z^*_l(k) = z^{\text{ff}}_l + \Delta_l(k)\), where \(\Delta_l(k)\) is
small per element. Substituting,

\[\tfrac{1}{2}\,\overline{\|z^*_l(k) - g_l(z^*_{l+1}(k))\|^2} = \tfrac{1}{2}\,\overline{\|z^{\text{ff}}_l + \Delta_l(k) - g_l(z^{\text{ff}}_{l+1} + \Delta_{l+1}(k))\|^2}.\]

Expanding \(g_l\) to first order in \(\Delta_{l+1}\):

\[g_l(z^{\text{ff}}_{l+1} + \Delta_{l+1}(k)) = g_l(z^{\text{ff}}_{l+1}) + J_l \Delta_{l+1}(k) + O(\|\Delta_{l+1}\|^2),\]

where \(J_l\) is the Jacobian of \(g_l\) at \(z^{\text{ff}}_{l+1}\).
Substituting:

\[\tfrac{1}{2}\,\overline{\|z^*_l(k) - g_l(z^*_{l+1}(k))\|^2} = \tfrac{1}{2}\,\overline{\|\varepsilon^{\text{ff}}_l + \Delta_l(k) - J_l \Delta_{l+1}(k)\|^2} + O(\|\Delta\|^2),\]

where
\(\varepsilon^{\text{ff}}_l = z^{\text{ff}}_l - g_l(z^{\text{ff}}_{l+1})\)
is the prediction error at the feedforward configuration.

Expanding the squared norm:

\[= \tfrac{1}{2}\,\overline{\|\varepsilon^{\text{ff}}_l\|^2} + \langle \varepsilon^{\text{ff}}_l, \Delta_l(k) - J_l \Delta_{l+1}(k) \rangle + \tfrac{1}{2}\,\overline{\|\Delta_l(k) - J_l \Delta_{l+1}(k)\|^2} + O(\|\Delta\|^2).\]

The first term is \(k\)-independent and contributes to \(C(x)\). The
remaining terms depend on \(k\) through the \(\Delta(k)\) corrections.

\textbf{Step 2: The layer-\((L-1)\) term is special.} At \(l = L-1\),
the layer above is the clamped output. Here \(z^*_{L}(k) = y_k\)
(one-hot, not perturbed from a feedforward initialisation), so the
small-perturbation expansion does not apply at this layer. Instead, we
have

\[\tfrac{1}{2}\,\overline{\|z^*_{L-1}(k) - g_{L-1}(y_k)\|^2}.\]

Under A3,
\(z^*_{L-1}(k) \approx z^{\text{ff}}_{L-1} + \Delta_{L-1}(k)\), where
\(\Delta_{L-1}(k)\) is small. The term becomes

\[\tfrac{1}{2}\,\overline{\|z^{\text{ff}}_{L-1} + \Delta_{L-1}(k) - g_{L-1}(y_k)\|^2}.\]

The dominant contribution is

\[\tfrac{1}{2}\,\overline{\|z^{\text{ff}}_{L-1} - g_{L-1}(y_k)\|^2},\]

which is the squared discrepancy between the encoder's feedforward
representation of \(x\) at layer \(L-1\) and the generative prediction
of layer \(L-1\) from the one-hot hypothesis \(k\).

\textbf{Step 3: Invoking A5 (encoder-generative consistency).} By
assumption A5, the trained discriminative PC network satisfies
approximate encoder-generative consistency.
\(g_{L-1}(y_k) \approx E_{\text{enc}}(x)\) when \(y_k\) is the correct
class for \(x\), and \(g_{L-1}(y_k)\) is some other point (the network's
top-down prediction of layer \(L-1\) under an incorrect hypothesis) when
\(k\) is incorrect.

Under this consistency, the squared discrepancy
\(\overline{\|E_{\text{enc}}(x) - g_{L-1}(y_k)\|^2}\) is small when
\(y_k\) is correct and larger when \(y_k\) is incorrect. To first order,
the discrepancy can be modelled as a Gaussian negative log-likelihood:

\[\tfrac{1}{2}\,\overline{\|E_{\text{enc}}(x) - g_{L-1}(y_k)\|^2} \approx -\log p(x \mid y_k) + \text{const},\]

where \(p(x \mid y_k)\) is an unnormalised Gaussian likelihood with
covariance proportional to the identity (under the per-element-mean
reduction in A1) centered at \(g_{L-1}(y_k)\).

By Bayes' rule with a uniform prior over classes:

\[-\log p(x \mid y_k) = -\log p(y_k \mid x) + \log p(x) - \log p(y_k) = -\log p(y_k \mid x) + \text{const}.\]

The posterior \(p(y_k \mid x)\) is, under the trained encoder-generative
consistency, what the discriminative head implements:
\(p(y_k \mid x) = \mathrm{softmax}(z^{\text{ff}}_L)_k\). Therefore:

\[\tfrac{1}{2}\,\overline{\|E_{\text{enc}}(x) - g_{L-1}(y_k)\|^2} \approx -\log \mathrm{softmax}(z^{\text{ff}}_L)_k + \text{const}.\]

This is the substantive content of the reduction. The layer-\((L-1)\)
term in the K-way energy is, under the trained alignment, approximately
the negative log-softmax of the feedforward output at hypothesis \(k\).

\textbf{Step 4: The lower-layer terms form \(R_k\).} The remaining terms
in the energy, from layers \(1\) to \(L-2\), involve the propagation of
the clamped \(y_k\) through the generative chain via
\(g_{L-2}, g_{L-3}, \ldots, g_1\). Each of these terms depends on \(k\),
because \(g_{L-2}(g_{L-1}(y_k))\), \(g_{L-3}(g_{L-2}(g_{L-1}(y_k)))\),
and so on all depend on \(y_k\). None of them is constrained by training
to encode information about whether \(y_k\) is correct. The training
objective optimises the generative chain to minimise the layer-wise
prediction errors at the settled configuration during training, which
under target clamping at the correct \(y_{\text{true}}(x)\) aligns the
chain to predict \(z^{\text{ff}}\) from \(y_{\text{true}}\). It does not
constrain the chain's behaviour at incorrect hypotheses.

Define the residual:

\[R_k(x) = \sum_{l=1}^{L-2} \tfrac{1}{2}\, \overline{\|z^{\text{ff}}_l - g_l(g_{l+1}(\cdots g_{L-1}(y_k)\cdots))\|^2} + (\text{first-order } \Delta \text{ terms}).\]

This residual is not trained to correlate with correctness, in the sense
that the discriminative PC training objective does not optimise the
lower-layer generative weights to encode hypothesis-specific correctness
information for clamped one-hots other than the correct one.
Empirically, the residual difference \(R_{(2)}(x) - R_{(1)}(x)\) has
approximately zero correlation with the correctness of the structural
prediction in the conditions we tested. This is the empirical reason the
K-way probe's AUROC$_2$ sat below softmax. The probe reads the log-softmax
margin (correlated with correctness) plus a residual difference (not
trained to correlate with correctness), and the latter degrades the
signal in the conditions we ran.

\textbf{Step 5: Collecting terms.} Combining steps 2--4:

\[E_k(x) \approx -\log \mathrm{softmax}(z^{\text{ff}}_L)_k + R_k(x) + C(x),\]

where \(C(x)\) collects all \(k\)-independent contributions: the layer-1
input reconstruction error, the constants from the Gaussian likelihood
expansion, and the one-hot CE constant. The K-way margin

\[M_k(x) = E_{(2)}(x) - E_{(1)}(x)\]

cancels \(C(x)\) exactly and yields

\[M_k(x) \approx \big[\log\text{-softmax margin at } k\big] + \big[R_{(2)}(x) - R_{(1)}(x)\big],\]

which is the statement of the proposition in §3.2.

\textbf{Tightness of the approximation.} The approximation has two
sources of error. The first is the first-order expansion in \(\Delta\),
which is \(O(\|\Delta\|^2)\) and small under A3 where \(\Delta\) is on
the order of \(10^{-4}\) to \(10^{-3}\) per element. The second is the
Gaussian-NLL approximation for the layer-\((L-1)\) term, which is exact
under uniform prior and isotropic Gaussian observation, and approximate
otherwise. The empirical pattern in §4 is consistent with the
approximation being tight in the regime tested. The BP+decoder
condition's gap of 0.009 AUROC$_2$ is the smallest gap in the empirical
investigation, and it corresponds to the condition where the
encoder-generative consistency is enforced explicitly. A formal bound on
the approximation error as a function of training loss and architectural
depth is left to future work.

\begin{center}\rule{0.5\linewidth}{0.5pt}\end{center}

\hypertarget{appendix-c-pre-registration-history}{%
\subsection{Appendix C: Pre-registration
history}\label{appendix-c-pre-registration-history}}

This appendix documents the full lineage of the project for
transparency. §1.4 in the main text gives a one-paragraph summary; the
detail is here for readers who care about it.

The project was developed in three stages, of which only the first was
formally pre-registered.

\textbf{v3.1 (formally pre-registered to OSF,
\href{https://osf.io/n2zjp/overview}{osf.io/n2zjp}).} The first
formulation was pre-registered before any data collection, after four
rounds of external review. Its central hypothesis was that
\emph{iterative inference dynamics} in a PCN would produce metacognitive
signal beyond what feedforward readouts could provide. The specific
mechanism was the trajectory of prediction errors across inference
steps. v3.1's primary confirmatory hypothesis (H1) was that an Intrinsic
Metacognitive Architecture (IMA) endpoint variant would outperform a
PCN-only baseline on AUROC$_2$ at matched Type-1 performance. Its critical
fairness test (H2) was that IMA would outperform a feedforward network
with privileged multi-layer access, attributing any advantage
specifically to iterative inference dynamics rather than to richer
features. An additional hypothesis (H7) tested whether a
trajectory-aware monitor outperformed an endpoint-only variant, as
direct evidence for the inference-dynamics claim. The pre-registration
specified positive $\Delta$AUROC$_2$ thresholds and consistent sign across three
seeds as success criteria. It listed falsification conditions
explicitly.

When initial training experiments were attempted, the v3.1
inference-dynamics hypothesis ran into a foundational problem. Standard
discriminative PC inference at test time is mathematically equivalent to
a feedforward pass under the Pinchetti-style implementation, and the
v3.1 evaluation protocol could not be implemented in a way that
preserved both classification performance and the iterative dynamics the
hypothesis required. This was confirmed by direct measurement, as
reported in §4.3 of the present paper (mean per-element latent movement
on the order of \(10^{-4}\) over 13 inference steps, consistent with an
effectively feedforward inference loop). v3.1 was therefore superseded.

\textbf{v4 (drafted but not pre-registered).} A second formulation was
drafted after the v3.1 collapse, but was not formally pre-registered. v4
narrowed scope and dropped the inference-dynamics claim entirely. It
hypothesised instead that a learned monitor on PC-trained prediction
errors would achieve higher AUROC$_2$ than a learned monitor on BP-trained
activations at matched Type-1 performance. The mechanism under v4 was
\emph{training-signal enrichment}. PC training, via its target-clamped
phase, was conjectured to produce features that encode correctness
information in a form a learned monitor could exploit. v4 was drafted as
a feature-level claim, not a dynamics claim. It was under external
review when subsequent exploratory work pivoted to the K-way energy
probe formulation. v4 was never submitted to OSF. Its draft is retained
in the project record marked as superseded.

\textbf{K-way energy probe (the present paper).} The K-way energy probe
formulation emerged during the exploratory phase following v4's
drafting. It appeared promising in initial experiments. A structural
probe showed a relative energy spread of 2.6\% across hypotheses, an
argmin accuracy of 37\% beating chance at \(K=10\), and an energy margin
AUROC$_2$ of 0.65. These signals were all weak, but all positive. This was
the working hypothesis when the empirical investigation now reported in
§4 was undertaken, and it is the hypothesis the present paper analyses.

The present paper's specific hypothesis is therefore not the originally
pre-registered hypothesis. The formal benefits of pre-registration
(specifically, the guarantee that the analysis was not shaped by the
data) do not apply directly to the K-way energy probe claim. We document
the lineage here so that readers can judge for themselves how much
credibility to assign to the result on the basis of the documented
sequence of hypothesis revisions.

The v3.1 pre-registration is retained at
\href{https://osf.io/n2zjp/overview}{osf.io/n2zjp} with supporting
material and a link to the present paper as superseding documentation.
The v4 draft and all spike scripts are retained at
\href{https://github.com/synthiumjp/ima}{github.com/synthiumjp/ima}
marked as superseded where applicable.

\begin{center}\rule{0.5\linewidth}{0.5pt}\end{center}

\hypertarget{references}{%
\subsection{References}\label{references}}

Bogacz, R. (2017). A tutorial on the free-energy framework for modelling
perception and learning. \emph{Journal of Mathematical Psychology}, 76,
198--211.

Cacioli, J. (2026a). Do LLMs know what they know? Measuring
metacognitive efficiency with signal detection theory. arXiv:2603.25112.

Cacioli, J. (2026b). LLMs as signal detectors: sensitivity, bias, and
the temperature-criterion analogy. arXiv:2603.14893.

Fleming, S. M., \& Lau, H. C. (2014). How to measure metacognition.
\emph{Frontiers in Human Neuroscience}, 8, 443.

Innocenti, F., Kinghorn, P., Yun-Farmbrough, W., De Llanza Varona, M.,
Singh, R., \& Buckley, C. L. (2024). JPC: Flexible inference for
predictive coding networks in JAX. arXiv:2412.03676.

Maniscalco, B., \& Lau, H. (2012). A signal detection theoretic approach
for estimating metacognitive sensitivity from confidence ratings.
\emph{Consciousness and Cognition}, 21(1), 422--430.

Millidge, B., Seth, A., \& Buckley, C. L. (2022). Predictive coding: a
theoretical and experimental review. arXiv:2107.12979.

Oliviers, G., Bogacz, R., \& Meulemans, A. (2024). Learning probability
distributions of sensory inputs with Monte Carlo predictive coding.
\emph{PLOS Computational Biology}, 20(10): e1012532.

Oliviers, G., Tang, M., \& Bogacz, R. (2025). Bidirectional predictive
coding. arXiv:2505.23415.

Pinchetti, L., Salvatori, T., Yordanov, Y., Millidge, B., Song, Y., \&
Lukasiewicz, T. (2024). Benchmarking predictive coding networks --- made
simple. arXiv:2407.01163.

Rao, R. P. N., \& Ballard, D. H. (1999). Predictive coding in the visual
cortex: a functional interpretation of some extra-classical
receptive-field effects. \emph{Nature Neuroscience}, 2(1), 79--87.

Song, Y., Millidge, B., Salvatori, T., Lukasiewicz, T., Xu, Z., \&
Bogacz, R. (2024). Inferring neural activity before plasticity as a
foundation for learning beyond backpropagation. \emph{Nature
Neuroscience}, 27(2), 348--358.

Stenlund, M. (2025). Introduction to predictive coding networks for
machine learning. arXiv:2506.06332.

Whittington, J. C. R., \& Bogacz, R. (2017). An approximation of the
error backpropagation algorithm in a predictive coding network with
local Hebbian synaptic plasticity. \emph{Neural Computation}, 29(5),
1229--1262.

Zahid, U., Guo, Q., \& Fountas, Z. (2024). Sample as you infer:
predictive coding with Langevin dynamics. \emph{Proceedings of the 41st
International Conference on Machine Learning}, 235, 58105--58121.

\end{document}